\documentclass{article}

\usepackage{PRIMEarxiv}

\usepackage[utf8]{inputenc} 
\usepackage[T1]{fontenc}    
\usepackage[round,authoryear]{natbib}
\usepackage{hyperref}       
\usepackage{url}            
\usepackage{booktabs}       
\usepackage{colortbl}       
\usepackage{float}          
\usepackage{amsmath}
\usepackage[ruled,vlined,linesnumbered]{algorithm2e}
\usepackage{tikz}
\usetikzlibrary{arrows.meta,decorations.pathreplacing}
\usepackage{amsfonts}       
\usepackage{mathcommands}    
\usepackage{nicefrac}       
\usepackage{microtype}      
\usepackage{fancyhdr}       
\usepackage{graphicx}       
\usepackage{enumitem}
\graphicspath{{media/}}     

\title{Bilateral Log-Linear Attention
}

\author{
  Kazi Ashik Islam\\
  Oak Ridge National Laboratory \\
  \texttt{islamk@ornl.gov} \\
   \And
  Guojing Cong \\
  Oak Ridge National Laboratory \\
  \texttt{congg@ornl.gov} \\
}

\begin{document}
\raggedbottom
\maketitle

\begin{abstract}
Sequence models represent past context at different resolutions, using token-level keys and values, fixed-size recurrent states, or hierarchies of compressed summaries. Full attention supports strong associative recall but requires quadratic work during training and prefill and a linearly growing key-value cache during inference. Linear attention and state-space models reduce these costs through fixed-size recurrent memory, but this compression can limit recall. Log-linear attention offers a middle ground by maintaining logarithmically many state matrices. However, its hierarchy favors finer resolution near the current query while compressing earlier context into coarser summaries. To complement this allocation of resolution, we introduce bilateral log-linear attention, which pairs the existing hierarchy with a second hierarchy anchored at the sequence origin. By combining their reads, the two branches provide finer access to both early and recent context while remaining causal. We derive recurrent and chunkwise parallel formulations for the added branch, preserving log-linear training and prefill work, logarithmic decoding work per token, and logarithmic persistent state storage at fixed feature dimensions and chunk size. In multi-query associative recall, this design raises mean overall test accuracy from 76.37\% to 96.12\% over forward log-linear attention across multiple training seeds. For pretrained language models under approximate parameter matching, it raises the mean score on six selected RULER needle-in-a-haystack tasks from 26.97\% to 50.03\% at a context length of 6144 tokens. In the same setting, bilateral attention also outperforms forward log-linear attention on seven of eight downstream retrieval tasks, including question answering and information extraction.
\end{abstract}

\section{Introduction}
\label{sec:introduction}

Attention enables Transformer models to make predictions using information distributed throughout the preceding context \citep{vaswani2017attention}. Its matrix formulation supports parallel computation across sequence positions, enabling efficient training on modern accelerators \citep{guo2026loglinear}. Despite this parallelism, the computation and memory storage required by dense attention grow with context length. At fixed feature dimensions, processing a sequence of length $T$ requires $O(T^2)$ arithmetic work. During autoregressive decoding, the key/value (KV) cache also grows linearly with the context. These costs become a bottleneck for longer sequences, particularly for inference on resource-constrained devices. They motivate context representations that support retrieval while limiting the growth of computation and storage.

An efficient context representation must support selective recall across the available sequence. This includes associative recall: retrieving information associated with a previously encountered cue. As contexts grow, more associations may need to be retained over longer distances. Compressing these associations into limited memory reduces storage, but can introduce interference that makes individual associations harder to retrieve. The challenge is therefore to balance compact storage with reliable recall across the context.

Linear attention \citep{katharopoulos2020transformers}, RetNet \citep{sun2023retentive}, H3~\citep{fu2023hungry}, RWKV~\citep{peng2023rwkv}, Gated Linear Attention \citep{yang2024gated}, and Mamba-2~\citep{dao2024transformers} reduce storage by compressing context into recurrent states whose size is independent of sequence length (see Section~\ref{sec:related_works} for a detailed literature review). Models in this family have been observed to underperform attention on associative recall~\citep{arora2024zoology,guo2026loglinear}. Log-linear attention~\citep{guo2026loglinear} provides a middle ground between fixed-size recurrent states and a Transformer's token-level KV cache. It maintains logarithmically many state matrices in a Fenwick hierarchy, grouping recent tokens into smaller buckets and older tokens into larger ones. This expands memory capacity while allocating finer context resolution near the query. Older associations remain represented, but more tokens share each state matrix (see Figure~\ref{fig:bg-context-resolution}). This recency bias motivates a complementary memory organization that supports recall across the context while preserving efficient sequence-length scaling.

We introduce \emph{bilateral log-linear attention}, a novel causal attention mechanism that combines two complementary hierarchies for efficient recall across the available context. It has two complementary branches: a forward branch that uses the Fenwick hierarchy of \citet{guo2026loglinear}, and a reverse branch whose hierarchy is anchored at the sequence origin, with bucket sizes growing geometrically toward later positions. Its finer grouping near the origin compensates for the Fenwick hierarchy's coarser grouping at the beginning of the context. The name \emph{reverse} refers to this opposite orientation of grouping; both branches access the same (past and current) tokens. 
The additional branch increases computation and persistent state storage by constant factors while preserving log-linear attention’s asymptotic scaling with sequence length. 

Specifically, our contributions are:

\begin{itemize}[leftmargin=*]
\item We present bilateral attention, which introduces a complementary causal hierarchy and integrates it with log-linear attention to support recall across the available context. It has finer grouping of tokens near the sequence origin and near the current query. Shared features, reverse write gating, branch normalization, and input-dependent branch mixing integrate the two memory organizations into a single attention mechanism.

\item We derive equivalent recurrent and matrix formulations of the reverse branch and develop a chunkwise evaluation algorithm that enables parallel execution through additive prefix scans. At fixed feature dimensions and chunk size, bilateral attention preserves log-linear attention’s asymptotic complexity: $O(T\log T)$ arithmetic work for training and prefill, $O(\log T)$ work per token during autoregressive decoding, and $O(\log T)$ persistent state storage.
\item Our evaluation on associative recall, language modeling, and retrieval compares bilateral attention with log-linear attention and Mamba-2, while using Transformer as a reference. On multi-query associative recall (MQAR) \citep{arora2024zoology}, bilateral attention achieves $96.12\%$ mean overall test accuracy versus $76.37\%$ for log-linear attention across ten training seeds, with depth and hidden width matched. On all six selected RULER needle-in-a-haystack tasks \citep{hsieh2024ruler}, bilateral attention outperforms log-linear attention at context lengths of 2048, 4096, and 6144 tokens. This holds under both layer matching and approximate parameter matching. At a context length of 6144 tokens, the approximately parameter-matched six-task mean improves from $26.97\%$ to $50.03\%$. Across eight downstream retrieval tasks, including three Natural Questions variants \citep{kwiatkowski2019natural}, bilateral attention outperforms log-linear attention on seven tasks under approximate parameter matching. 
These results narrow the retrieval-performance gap to Transformer on the evaluated benchmarks while retaining logarithmic persistent state storage.
\end{itemize}

\section{Background}
\label{sec:background}

We consider causal sequence modeling, where the output at position $t$ depends only on positions $0,\ldots,t$. For clarity, we describe one attention head and omit batch and layer indices. Let $\bq_t,\bk_t,\bv_t,\bo_t\in\mathbb R^d$ denote the query, key, value, and output vectors, and let $\Qb,\Kb,\Vb,\Ob\in\mathbb R^{T\times d}$ stack their transposes as rows. Following \citet{guo2026loglinear}, we distinguish two equivalent views of efficient attention: ($i$) a matrix form that exposes parallel computation across a known sequence during training, and ($ii$) a recurrent form that summarizes the prefix during autoregressive inference. 

\subsection{Linear attention and scalar decay}
\label{sec:background-linear}

Standard softmax attention retains a key and value for each past token, requiring $O(T)$ cache storage (referred to as the KV cache) during decoding and $O(T^2)$ time to process a length-$T$ sequence at fixed embedding dimension. Linear attention instead utilizes the associativity of query, key, and value matrix products to summarize the past tokens in a matrix state \citep{katharopoulos2020transformers}. We use the unnormalized dot-product formulation adopted by \citet{guo2026loglinear}. 
Its matrix form is:
\begin{equation}
\Ob=(\Qb\Kb^\top\odot \Mb)\Vb,
\qquad
\Mb_{ts}=\begin{cases}1,&s\le t,\\0,&s>t,\end{cases}
\label{eq:bg-linear-matrix}
\end{equation}
Here, $\odot$ denotes elementwise multiplication. The lower-triangular mask ($\Mb$) enforces causality, including the current position. 
The corresponding recurrent form is:
\begin{equation}
\Sbb_{0}=\bv_0\bk_0^\top,
\qquad \Sbb_t=\Sbb_{t-1}+\bv_t\bk_t^\top,
\qquad \bo_t=\Sbb_t\bq_t, \qquad \forall t > 0
\label{eq:bg-linear-recurrence}
\end{equation}
where $\Sbb_t\in\mathbb R^{d\times d}$. Each key/value association contributes an
outer product, and the query reads their accumulated summary. \cite{dao2024transformers} extended this formulation in Mamba-2 by introducing a data-dependent scalar decay.
\begin{equation}
\Sbb_{0}=\bv_0\bk_0^\top,
\qquad \Sbb_t=\alpha_t \Sbb_{t-1}+\bv_t\bk_t^\top,
\qquad \bo_t=\Sbb_t\bq_t, \qquad \forall t > 0
\label{eq:bg-mamba2-recurrence}
\end{equation}
A scalar $\alpha_t\in(0,1)$ attenuates the previous state before inserting the current association. Setting $\alpha_t=1$ recovers ordinary linear attention in Eq.~\eqref{eq:bg-linear-recurrence}. 
Unrolling Eq.~\eqref{eq:bg-mamba2-recurrence} gives the matrix form:
\begin{equation}
\Ob=(\Qb\Kb^\top\odot \Mb^S)\Vb,
\qquad
\Mb^S_{ts}=\begin{cases}
\displaystyle\prod_{j=s+1}^{t}\alpha_j,&s\le t,\\
0,&s>t.
\end{cases}
\label{eq:bg-decay-matrix}
\end{equation}
The matrix and recurrent forms describe the same operation. However, their computational costs differ. Efficient chunkwise algorithms exploit the temporal structure in $\Mb^S$ to evaluate the sequence in parallel with $O(T)$ time complexity at fixed embedding dimensions and chunk size. Directly materializing the matrix in Eq.~\eqref{eq:bg-decay-matrix} would still incur quadratic complexity; the efficient bound follows from its structure, rather than from the absence of a softmax alone. The recurrent form, used during decoding, updates and reads a single state with $O(1)$ time and memory per token. Direct evaluation of the recurrence processes tokens sequentially; parallel training instead uses an equivalent scan-based or chunkwise formulation.

Figure~\ref{fig:bg-context-resolution}(a) illustrates the resulting memory representation. Ordinary linear attention and its scalar-decay variant both compress all past associations into one state matrix. Decay changes the contributions retained in this state, but it does not increase the number of independently accessible temporal summaries. 

\begingroup
\definecolor{bgSingle}{HTML}{DCE8F1}
\definecolor{bgEight}{HTML}{FCE2BF}
\definecolor{bgFour}{HTML}{F6C27C}
\definecolor{bgTwo}{HTML}{D99535}
\definecolor{bgOne}{HTML}{955B16}
\newcommand{\bgmatrix}[4]{%
  \begin{scope}[shift={(#1,#2)}]
    \filldraw[fill=#3,draw=black!65,line width=0.6pt] (-0.55,-0.55) rectangle (0.55,0.55);
    \foreach \g in {-0.275,0,0.275}{
      \draw[black!15,line width=0.3pt] (\g,-0.55)--(\g,0.55);
      \draw[black!15,line width=0.3pt] (-0.55,\g)--(0.55,\g);
    }
    \node[fill=#3,inner sep=1.5pt] at (0,0) {$#4$};
  \end{scope}%
}
\begin{figure}[tbp]
\centering
\resizebox{0.85\linewidth}{!}{%
\begin{tikzpicture}[x=0.7cm,y=0.7cm,font=\small,>=Stealth]
  \node[anchor=west,font=\small\bfseries] at (0,9.0) {(a) Linear attention / Mamba-2};
  \node[anchor=east,font=\footnotesize] at (15,9.0) {$t=14$ (15 tokens)};
  \node[anchor=west,font=\footnotesize,text=black!65] at (0,8.45) {older tokens};
  \node[anchor=east,font=\footnotesize,text=black!65] at (15,8.45) {recent tokens};
  \foreach \i in {0,...,14}{
    \filldraw[fill=bgSingle,draw=black!35,line width=0.3pt] (\i,7.5) rectangle (\i+1,8.1);
    \node at (\i+0.5,7.8) {\i};
  }
  \draw[black!70,line width=0.8pt] (0,7.5) rectangle (15,8.1);
  \draw[->,black!70] (7.5,7.5)--(7.5,6.85);
  \bgmatrix{7.5}{6.25}{bgSingle}{\Sbb_t}
  \node[anchor=west,align=left,font=\footnotesize] at (0,6.25) {one $d\times d$ state per head\\$O(1)$ recurrent memory};
  \draw[->,black!70] (8.1,6.25)--(10.6,6.25);
  \node[anchor=west] at (10.8,6.25) {$\bo_t=\Sbb_t\bq_t$};
  \node[font=\footnotesize,text=black!65] at (7.5,5.35) {all associations share one summary};
  \draw[black!20] (0,4.95)--(15,4.95);

  \node[anchor=west,font=\small\bfseries] at (0,4.5) {(b) Log-linear attention};
  \foreach \i in {0,...,7}{
    \filldraw[fill=bgEight,draw=black!30,line width=0.3pt] (\i,3.4) rectangle (\i+1,4.0);
    \node at (\i+0.5,3.7) {\i};
  }
  \foreach \i in {8,...,11}{
    \filldraw[fill=bgFour,draw=black!30,line width=0.3pt] (\i,3.4) rectangle (\i+1,4.0);
    \node at (\i+0.5,3.7) {\i};
  }
  \foreach \i in {12,13}{
    \filldraw[fill=bgTwo,draw=black!30,line width=0.3pt] (\i,3.4) rectangle (\i+1,4.0);
    \node at (\i+0.5,3.7) {\i};
  }
  \filldraw[fill=bgOne,draw=black!65,line width=0.6pt] (14,3.4) rectangle (15,4.0);
  \node[text=white] at (14.5,3.7) {14};
  \foreach \a/\b in {0/8,8/12,12/14,14/15}{
    \draw[black!70,line width=0.8pt] (\a,3.4) rectangle (\b,4.0);
  }
  \foreach \a/\b/\ell in {0/8/4,8/12/3,12/14/2,14/15/0}{
    \draw[decorate,decoration={brace,mirror,amplitude=3pt},draw=black!70,line width=0.4pt]
      (\a+0.06,3.25)--(\b-0.06,3.25)
      node[midway,below=5pt,font=\footnotesize,inner sep=1pt] (bgBucket\ell) {$\cB_t^{(\ell)}$};
  }
  \foreach \x/\lev in {4/4,10/3,13/2,14.5/0}{
    \draw[->,black!70] ([yshift=-3pt]bgBucket\lev.south)--(\x,1.55);
  }
  \bgmatrix{4}{0.95}{bgEight}{\Sbb_t^{(4)}}
  \bgmatrix{10}{0.95}{bgFour}{\Sbb_t^{(3)}}
  \bgmatrix{13}{0.95}{bgTwo}{\Sbb_t^{(2)}}
  \bgmatrix{14.5}{0.95}{bgOne}{\color{white}\Sbb_t^{(0)}}
  \node[anchor=west,align=left,font=\footnotesize] at (0,0.95) {$O(\log T)$ states\\of the same size};
  \foreach \x/\ell in {4/4,10/3,13/2,14.5/0}{
    \draw[black!70] (\x,0.35)--(\x,-0.4);
    \node[fill=white,inner sep=1pt,font=\footnotesize] at (\x,-0.05) {$\lambda_t^{(\ell)}$};
  }
  \draw[black!70] (4,-0.4)--(14.5,-0.4);
  \draw[->,black!70] (9,-0.4)--(9,-0.9);
  \node at (9,-1.3) {$\bo_t=\sum_{\ell}\lambda_t^{(\ell)}\Sbb_t^{(\ell)}\bq_t$};
  \node[anchor=west,font=\footnotesize,text=black!65] at (0,-2.0) {coarser summaries for older tokens};
  \node[anchor=east,font=\footnotesize,text=black!65] at (15,-2.0) {finer summaries near the query};
\end{tikzpicture}%
}
\caption{\textbf{Context representation in Linear attention / Mamba-2 and Log-linear attention.} At $t=14$, the causal prefix contains 15 tokens. (a) Linear attention and Mamba-2 summarize the entire prefix in one state matrix per head; scalar decay changes its contents without creating separate temporal buckets. (b) Log-linear attention groups tokens into buckets of sizes $8,4,2,1$, labeled by braces as $\cB_t^{(\ell)}$. Each bucket is represented by a separate state matrix of the same dimension and is read with its own level weight $\lambda_t^{(\ell)}$. The nonempty levels are $4,3,2,0$; level 1 is empty ($\Sbb_t^{(1)} = \zerob$). Color indicates bucket size, not attention strength. 
The hierarchical construction follows \citet{guo2026loglinear}.}
\label{fig:bg-context-resolution}
\end{figure}
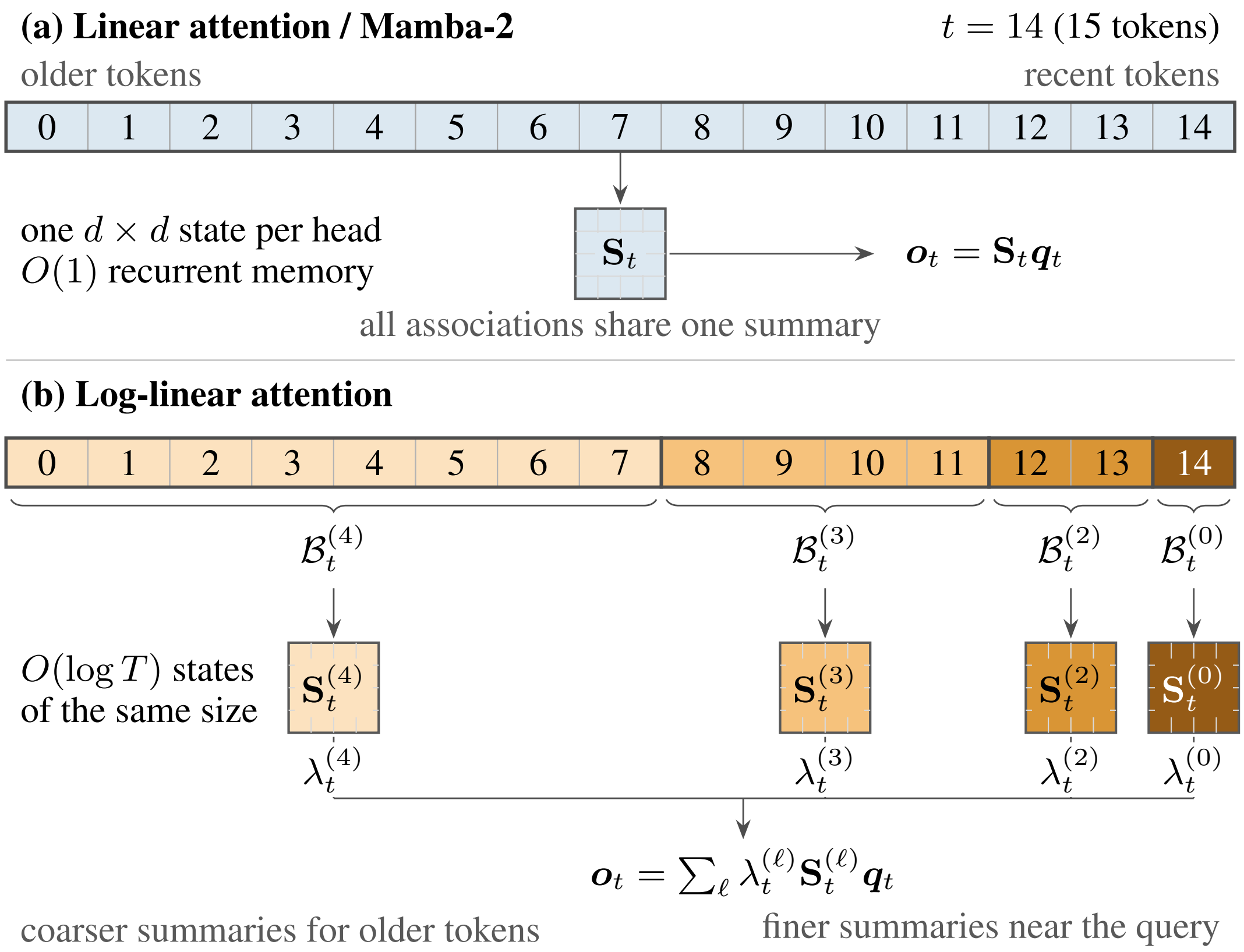
\endgroup

\subsection{Log-linear attention}
\label{sec:background-loglinear}
Log-linear attention \citep{guo2026loglinear} occupies an intermediate point between a single state matrix summary and a token-level KV cache. It partitions the causal prefix using a Fenwick hierarchy and retains one state matrix for each nonempty bucket of tokens (see Figure~\ref{fig:bg-context-resolution}(b)). Let $\mathcal B_t^{(\ell)}$ denote the bucket at level $\ell$ for query position $t$. Level zero contains the current token. Each nonempty bucket at a historical level $(\ell\ge1)$ contains $(2^{\ell-1})$ consecutive earlier tokens. These buckets form a disjoint partition of $\{0,\ldots,t\}$, with $O(\log T)$ levels sufficient for a sequence of length $T$. Specifically, $L=\lceil\log_2 T\rceil+1$ levels suffice, including level zero. 

In the undecayed case, the state representation and query read are:
\begin{equation}
\Sbb_t^{(\ell)}=\sum_{s\in\mathcal B_t^{(\ell)}}\bv_s\bk_s^\top,
\qquad
\bo_t=\sum_{\ell=0}^{L-1}\lambda_t^{(\ell)}\Sbb_t^{(\ell)}\bq_t,
\label{eq:bg-loglinear-recurrent}
\end{equation}
where empty levels contribute zero and the nonnegative level weights $\lambda_t^{(\ell)}$ depend on the current token. These level weights allow the model to do a weighted selection of temporal summaries at read time; they need not sum to one. The states admit online updates through the Fenwick carry/merge rule, rather than requiring the defining sums to be recomputed from all past tokens. We refer to \citet[Section~3.2]{guo2026loglinear} for the detailed recurrence.

The equivalent matrix form is:
\begin{equation}
\Ob=(\Qb\Kb^\top\odot \Mb^H)\Vb,
\qquad
\Mb^H_{ts}=\begin{cases}
\lambda_t^{(\ell(t,s))},&s\le t,\\
0,&s>t,
\end{cases}
\label{eq:bg-loglinear-matrix}
\end{equation}
where $\ell(t,s)$ identifies the bucket containing source $s$ at query $t$. Thus each source in the same bucket receives the same level weight, in addition to its query/key interaction. 

With scalar decay, referred to as `Log-linear Mamba-2' in~\cite{guo2026loglinear}, the matrix form is:
\begin{equation}
\Ob=(\Qb\Kb^\top\odot \Mb^S\odot \Mb^H)\Vb.
\label{eq:bg-loglinear-decay}
\end{equation}
Here, $\Mb^S$ is the mask matrix from Eq.~\eqref{eq:bg-decay-matrix}. In recurrent form, this replaces $\Sbb_t^{(\ell)}$ by $\sum_{s\in\mathcal B_t^{(\ell)}}\Mb^S_{ts}\bv_s\bk_s^\top$, leaving the read in Eq.~\eqref{eq:bg-loglinear-recurrent} unchanged. 

The hierarchy exposes different parts of the prefix at different resolutions. For the 15-token example in Figure~\ref{fig:bg-context-resolution}(b), its nonempty buckets are $\{0,\ldots,7\}$, $\{8,\ldots,11\}$, $\{12,13\}$, and $\{14\}$. Older positions share coarser summaries, while recent positions occupy smaller buckets. Here, resolution refers to the grouping of sources into independently weighted matrix states.

The matrix and recurrent views also make the computational tradeoff explicit. The hierarchical structure supports chunkwise parallel training with $O(T\log T)$ time complexity, while online decoding reads and maintains $O(\log T)$ matrix states. Table~\ref{tab:bg-complexity} summarizes the sequence-length scaling. These decoding storage bounds are for persistent states only.

\begin{table}[tbp]
\centering
\small
\begin{tabular}{lccc}
\toprule
Mechanism & Training time & Decode time per token & Decode memory \\
\midrule
Softmax attention & $O(T^2)$ & $O(T)$ & $O(T)$ \\
Linear attention w/wo. scalar decay & $O(T)$ & $O(1)$ & $O(1)$ \\
Log-linear attention w/wo. scalar decay & $O(T\log T)$ & $O(\log T)$ & $O(\log T)$ \\
\bottomrule
\end{tabular}
\vspace{2mm}
\caption{Complexity in terms of sequence-length at fixed model dimensions and chunk size, following \citet{guo2026loglinear}. Training time refers to complexity over a length-$T$ sequence; decode work is for one new token. Decode memory refers to persistent recurrent state or KV cache storage.}
\label{tab:bg-complexity}
\end{table}

The Fenwick hierarchy increases the number of temporal summaries, while favoring finer grouping near the query. Earlier positions remain represented, but are aggregated into progressively coarser states. In the scalar-decay variant, their contributions are also attenuated by subsequent decay factors. This motivates supplementing the Fenwick hierarchy with a complementary causal partition that allocates context resolution differently while preserving logarithmic recurrent storage. The next section introduces bilateral log-linear attention, which combines the Fenwick hierarchy with a second hierarchy anchored at the sequence origin.

\section{Bilateral Log-Linear Attention}
\label{sec:bilateral-loglinear}

Bilateral log-linear attention augments the Fenwick hierarchy of \citet{guo2026loglinear} with a second hierarchy that allocates fine context resolution in the opposite direction. The two hierarchies/branches organize the same causal prefix into different sets of matrix summaries, using shared query, key, and base value features. We call the inherited Fenwick view the \emph{forward hierarchy/branch} and the complementary, origin-anchored view the \emph{reverse hierarchy/branch}. Both branches access only past and current tokens; reverse refers to the orientation of the grouping, rather than a backward pass over the sequence.

\subsection{Complementary context resolution}
\label{sec:bilateral-resolution}

The forward hierarchy groups recent tokens into smaller buckets and older tokens into larger ones, as illustrated in Figure~\ref{fig:bg-context-resolution}(b). When many tokens share a state matrix, their stored associations can interfere, making it harder to retrieve information about individual tokens. This motivates a complementary hierarchy with smaller buckets near the sequence origin, providing finer access to early context. We let bucket sizes grow geometrically for later positions to keep the number of states logarithmic in the sequence length. Together, the two hierarchies provide finer access to both recent and early context.

Figure~\ref{fig:bilateral-context-resolution} compares the two views at $t=14$. The forward branch groups the 15 visible positions into $\{0,\ldots,7\}$, $\{8,\ldots,11\}$, $\{12,13\}$, and $\{14\}$. The reverse branch instead groups them into $\{0\}$, $\{1,2\}$, $\{3,\ldots,6\}$, and $\{7,\ldots,14\}$. For example, position zero has its own reverse summary, whereas it shares a forward summary with positions one through seven. Near the query, the forward branch provides the finer grouping. Thus the second hierarchy complements the existing view, without changing which source positions are causally available.

Each bucket is represented by a matrix of fixed dimensions, regardless of its number of tokens. Here, finer resolution means that fewer tokens share an independently weighted summary; it does not imply lossless recovery of the associations within that summary. In this example, each branch has four nonempty states, although the number of populated forward states varies with the query position. The geometric reverse buckets require only logarithmically many states as the prefix grows.

\begingroup
\definecolor{bilEight}{HTML}{FCE2BF}
\definecolor{bilFour}{HTML}{F6C27C}
\definecolor{bilTwo}{HTML}{D99535}
\definecolor{bilOne}{HTML}{955B16}
\newcommand{\bilstate}[4]{%
 \begin{scope}[shift={(#1,#2)}]
  \filldraw[fill=#3,draw=black!65,line width=0.6pt] (-0.55,-0.55) rectangle (0.55,0.55);
  \foreach \g in {-0.275,0,0.275}{
   \draw[black!15,line width=0.3pt] (\g,-0.55)--(\g,0.55);
   \draw[black!15,line width=0.3pt] (-0.55,\g)--(0.55,\g);
  }
  \node[fill=#3,inner sep=1pt] at (0,0) {$\scriptstyle #4$};
 \end{scope}%
}
\begin{figure}[tbp]
\centering
\resizebox{0.85\linewidth}{!}{%
\begin{tikzpicture}[x=0.7cm,y=0.7cm,font=\small,>=Stealth]
 \node[anchor=west,font=\small\bfseries] at (0,7.9) {(a) Forward hierarchy};
 \node[anchor=east,font=\footnotesize] at (15,7.9) {$t=14$ (15 tokens)};
 \node[anchor=west,font=\footnotesize,text=black!65] at (0,7.35) {older tokens};
 \node[anchor=east,font=\footnotesize,text=black!65] at (15,7.35) {recent tokens};
 \foreach \i in {0,...,7}{
  \filldraw[fill=bilEight,draw=black!30,line width=0.3pt] (\i,6.3) rectangle (\i+1,6.9);
  \node at (\i+0.5,6.6) {\i};
 }
 \foreach \i in {8,...,11}{
  \filldraw[fill=bilFour,draw=black!30,line width=0.3pt] (\i,6.3) rectangle (\i+1,6.9);
  \node at (\i+0.5,6.6) {\i};
 }
 \foreach \i in {12,13}{
  \filldraw[fill=bilTwo,draw=black!30,line width=0.3pt] (\i,6.3) rectangle (\i+1,6.9);
  \node at (\i+0.5,6.6) {\i};
 }
 \filldraw[fill=bilOne,draw=black!65,line width=0.6pt] (14,6.3) rectangle (15,6.9);
 \node[text=white] at (14.5,6.6) {14};
 \foreach \a/\b in {0/8,8/12,12/14,14/15}{\draw[black!70,line width=0.8pt] (\a,6.3) rectangle (\b,6.9);}
 \foreach \a/\b/\ell in {0/8/4,8/12/3,12/14/2,14/15/0}{
  \draw[decorate,decoration={brace,mirror,amplitude=3pt},draw=black!70,line width=0.4pt]
   (\a+0.06,6.15)--(\b-0.06,6.15)
   node[midway,below=5pt,font=\footnotesize,inner sep=1pt] (bilFBucket\ell) {$\cB_t^{\mathrm f,(\ell)}$};
 }
 \foreach \x/\ell in {4/4,10/3,13/2,14.5/0}{
  \draw[->,black!70] ([yshift=-3pt]bilFBucket\ell.south)--(\x,4.75);
 }
 \bilstate{4}{4.15}{bilEight}{\Sbb_t^{\mathrm f,(4)}}
 \bilstate{10}{4.15}{bilFour}{\Sbb_t^{\mathrm f,(3)}}
 \bilstate{13}{4.15}{bilTwo}{\Sbb_t^{\mathrm f,(2)}}
 \bilstate{14.5}{4.15}{bilOne}{\color{white}\Sbb_t^{\mathrm f,(0)}}
 \foreach \x/\ell in {4/4,10/3,13/2,14.5/0}{\node[font=\footnotesize] at (\x,3.25) {$\lambda_t^{(\ell)}$};}
 \node[anchor=west,align=left,font=\footnotesize,text=black!65] at (0,4.15) {coarser early\\summaries};
 \draw[black!20] (0,2.65)--(15,2.65);
 \node[anchor=west,font=\small\bfseries] at (0,2.15) {(b) Reverse hierarchy};
 \filldraw[fill=bilOne,draw=black!65,line width=0.6pt] (0,0.9) rectangle (1,1.5);
 \node[text=white] at (0.5,1.2) {0};
 \foreach \i in {1,2}{
  \filldraw[fill=bilTwo,draw=black!30,line width=0.3pt] (\i,0.9) rectangle (\i+1,1.5);
  \node at (\i+0.5,1.2) {\i};
 }
 \foreach \i in {3,...,6}{
  \filldraw[fill=bilFour,draw=black!30,line width=0.3pt] (\i,0.9) rectangle (\i+1,1.5);
  \node at (\i+0.5,1.2) {\i};
 }
 \foreach \i in {7,...,14}{
  \filldraw[fill=bilEight,draw=black!30,line width=0.3pt] (\i,0.9) rectangle (\i+1,1.5);
  \node at (\i+0.5,1.2) {\i};
 }
 \foreach \a/\b in {0/1,1/3,3/7,7/15}{\draw[black!70,line width=0.8pt] (\a,0.9) rectangle (\b,1.5);}
 \foreach \a/\b/\ell in {0/1/0,1/3/1,3/7/2,7/15/3}{
  \draw[decorate,decoration={brace,mirror,amplitude=3pt},draw=black!70,line width=0.4pt]
   (\a+0.06,0.75)--(\b-0.06,0.75)
   node[midway,below=5pt,font=\footnotesize,inner sep=1pt] (bilRBucket\ell) {$\cB_t^{\mathrm r,(\ell)}$};
 }
 \foreach \x/\ell in {0.5/0,2/1,5/2,11/3}{
  \draw[->,black!70] ([yshift=-3pt]bilRBucket\ell.south)--(\x,-0.65);
 }
 \bilstate{0.5}{-1.25}{bilOne}{\color{white}\Sbb_t^{\mathrm r,(0)}}
 \bilstate{2}{-1.25}{bilTwo}{\Sbb_t^{\mathrm r,(1)}}
 \bilstate{5}{-1.25}{bilFour}{\Sbb_t^{\mathrm r,(2)}}
 \bilstate{11}{-1.25}{bilEight}{\Sbb_t^{\mathrm r,(3)}}
 \node[font=\footnotesize] at (0.5,-2.2) {$\mu_t^{(0)}$};
 \node[font=\footnotesize] at (2,-2.2) {$\mu_t^{(1)}$};
 \node[font=\footnotesize] at (5,-2.2) {$\mu_t^{(2)}$};
 \node[font=\footnotesize] at (11,-2.2) {$\mu_t^{(3)}$};
 \node[anchor=east,align=right,font=\footnotesize,text=black!65] at (15,-1.25) {coarser recent\\summaries};
\end{tikzpicture}%
}
\caption{\textbf{Complementary context resolution.} At $t=14$, the forward Fenwick view groups 15 positions into buckets of sizes $8,4,2,1$, while the stationary reverse view uses sizes $1,2,4,8$. Braces label the visible buckets $\cB_t^{\mathrm f,(\ell)}$ and $\cB_t^{\mathrm r,(\ell)}$. Each bucket feeds one matrix of the same dimensions, read with the current query and the indicated level weight. Reverse states accumulate write-gated associations. Colors encode bucket size, consistently with Figure~\ref{fig:bg-context-resolution}. Both views are causal and store compressed summaries. Forward level 1 is empty in this example.}
\label{fig:bilateral-context-resolution}
\end{figure}
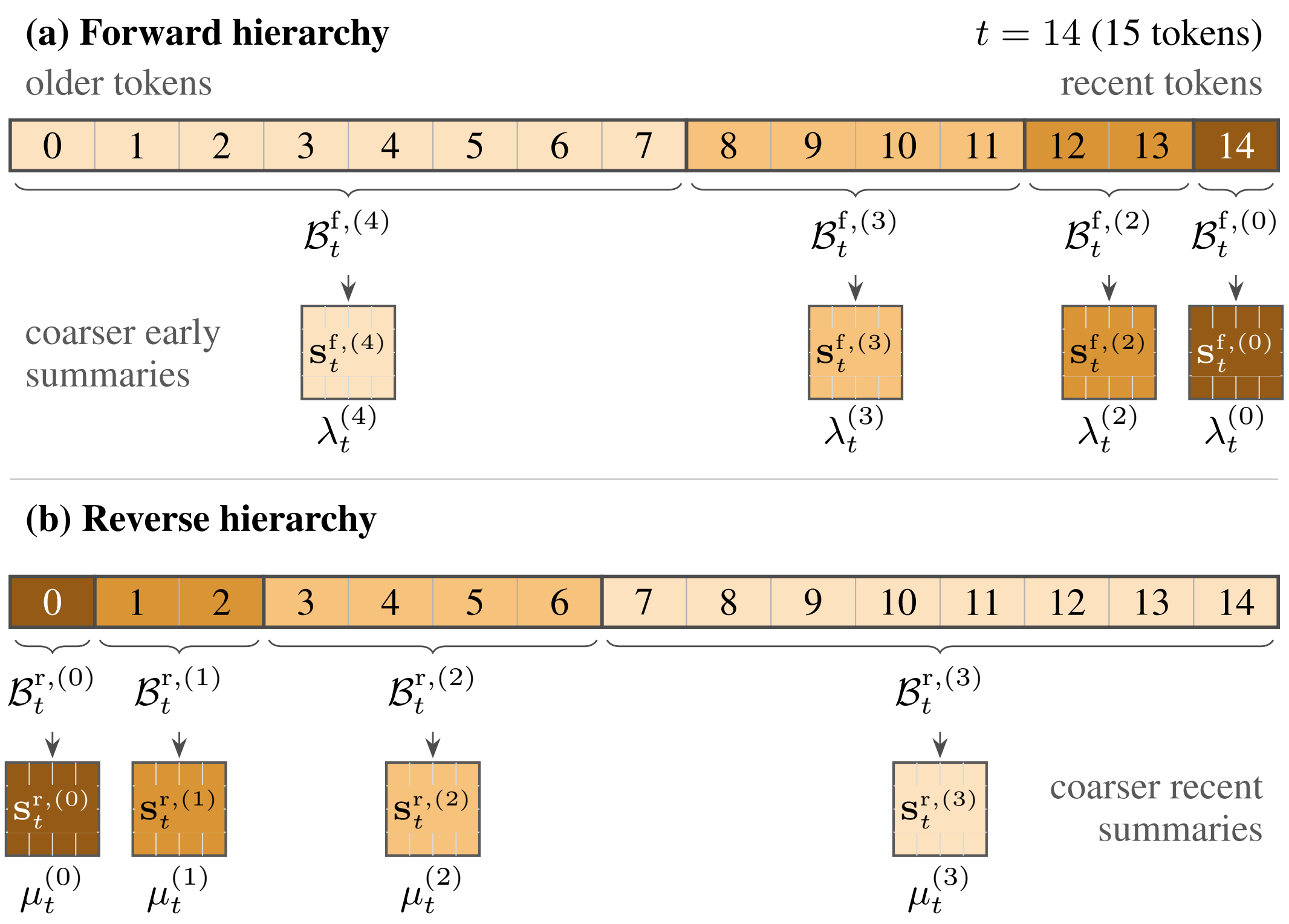
\endgroup

\subsection{Reverse branch attention}
\label{sec:reverse-branch-attention}

We retain the column-vector convention of Section~\ref{sec:background} and suppress batch, layer, and head indices. To allow different key and value dimensions, let $\bq_t,\bk_t\in\RR^{d_k}$ and $\bv_t\in\RR^{d_v}$; Section~\ref{sec:background} uses the special case $d_k=d_v=d$. The reverse state matrices therefore have shape $d_v\times d_k$.

A source position $s$ is assigned to the fixed level
\begin{equation}
\ell_{\mathrm r}(s)=\left\lfloor\log_2(s+1)\right\rfloor.
\label{eq:reverse-level}
\end{equation}
The full bucket at level $\ell$ and its visible portion at query $t$ are
\begin{equation}
\cC_\ell=\{2^\ell-1,\ldots,2^{\ell+1}-2\},
\qquad
\cB_t^{\mathrm r,(\ell)}=\cC_\ell\cap\{0,\ldots,t\}.
\label{eq:reverse-buckets}
\end{equation}
The full bucket contains $2^\ell$ positions. Unlike the query-dependent Fenwick assignment, a source's reverse level remains fixed as the query advances. The visible buckets form a disjoint partition of the causal prefix, and the intersection in Eq.~\eqref{eq:reverse-buckets} excludes any future positions in an unfinished bucket. 

For a sequence of length $T\ge1$, the reverse hierarchy has $L_{\mathrm r}=\lfloor\log_2 T\rfloor+1$ nonempty buckets, with levels $0,\ldots,L_{\mathrm r}-1$. The final bucket may be partially filled.

\subsubsection{Recurrent form}
\label{sec:reverse-recurrent}

For autoregressive decoding, let $\eta_s$ denote the data-dependent scalar write gate for source token $s$. Each visible bucket is represented by the state matrix:
\begin{equation}
\Sbb_t^{\mathrm r,(\ell)}
 =\sum_{s\in\cB_t^{\mathrm r,(\ell)}}\eta_s\bv_s\bk_s^\top.
\label{eq:reverse-state-definition}
\end{equation}
Empty buckets have zero state. Initialize $\Sbb_{-1}^{\mathrm r,(\ell)}=\zerob$ for every level. Since the current token belongs to level $\ell_{\mathrm r}(t)$ (Eq.~\eqref{eq:reverse-level}), the online update is
\begin{equation}
\Sbb_t^{\mathrm r,(\ell)}
 =\Sbb_{t-1}^{\mathrm r,(\ell)}
  +\mathbf 1\{\ell=\ell_{\mathrm r}(t)\}\,\eta_t\bv_t\bk_t^\top.
\label{eq:reverse-recurrence}
\end{equation}
The write gate lets the model control how strongly each incoming association is stored. It can suppress less useful writes without repeatedly attenuating associations already in memory. The gate acts only on the reverse write; i.e., it does not change the forward branch computation.

Only one level's state matrix is updated using the new gated outer product. All other reverse state matrices remain unchanged. There is no reverse carry, merge, or temporal decay; this makes the recurrent form of the reverse branch simpler than the forward branch. For example, token 14 completes level 3, and token 15 begins level 4 while retaining the completed lower-level summaries. Equation~\eqref{eq:reverse-recurrence} maintains the bucket sum in Eq.~\eqref{eq:reverse-state-definition}: every earlier token retains its assignment and gated contribution, and the only new token is $t$.

The current query reads the updated states as:
\begin{equation}
\bo_t^{\mathrm r}
 =\sum_{\ell=0}^{L_{\mathrm r}-1}\mu_t^{(\ell)}
   \Sbb_t^{\mathrm r,(\ell)}\bq_t.
\label{eq:reverse-state-read}
\end{equation}
The write gate $\eta_s$ controls how strongly source token $s$ contributes to its stored bucket summary. The level weight $\mu_t^{(\ell)}$ controls how strongly query $t$ reads that summary. We apply the write gate once when the token enters memory, whereas the level weight is recomputed for each query. Thus the state contains gated source contributions, while later queries can weight the same summary differently.

The positive, input-dependent level weights $\mu_t^{(\ell)}$ are parameterized independently of the forward level weights $\lambda_t^{(\ell)}$. They need not sum to one. Insertion precedes reading, and hence the current token contributes to its own output.

The reverse branch accumulates write-gated associations without temporal decay. Once a bucket is full, its stored matrix remains unchanged, while its query and level weight remain adaptive. This preserves the summary, although tokens sharing a finite state matrix can still interfere.

\subsubsection{Matrix Form and Efficient Computation}
\label{sec:reverse-parallel}

The operation described in Section~\ref{sec:reverse-recurrent} can be written over an entire sequence. Let $\bm{\eta}=(\eta_0,\ldots,\eta_{T-1})^\top$ denote the source write gates for tokens $0, ..., T-1$. Using the matrix convention of Eq.~\eqref{eq:bg-loglinear-matrix}, define
\begin{equation}
\Ob^{\mathrm r}=(\Qb\Kb^\top\odot\Mb^{H,\mathrm r})\operatorname{diag}(\bm{\eta})\Vb,
\qquad
\Mb^{H,\mathrm r}_{ts}
 =\begin{cases}
 \mu_t^{(\ell_{\mathrm r}(s))},&s\le t,\\
 0,&s>t.
 \end{cases}
\label{eq:reverse-matrix}
\end{equation}
Here $\Qb,\Kb\in\RR^{T\times d_k}$ and $\Vb,\Ob^{\mathrm r}\in\RR^{T\times d_v}$. Row $t$ of $\Ob^{\mathrm r}$ is the transpose of Eq.~\eqref{eq:reverse-state-read}: source $s\le t$ contributes $\mu_t^{(\ell_{\mathrm r}(s))}\eta_s(\bq_t^\top\bk_s)\bv_s$ in both forms. Thus the matrix form and the recurrence describe the same causal operator, including the current token. The diagonal matrix scales source values before accumulation and need not be materialized. It leaves the hierarchical mask unchanged.

For the forward branch, we use the efficient chunkwise algorithm of \citet[Section~3.3]{guo2026loglinear}. For each reverse bucket, Figure~\ref{fig:reverse-mask-structure} separates within-bucket causal reads (the triangle) from after-bucket reads of its completed summary (the rectangle). Queries before the bucket receive no contribution from that bucket.

\begingroup
\definecolor{revEight}{HTML}{AA4499}
\definecolor{revFour}{HTML}{009E73}
\definecolor{revTwo}{HTML}{D55E00}
\definecolor{revOne}{HTML}{0072B2}
\colorlet{revOneEdge}{revOne}
\colorlet{revTwoEdge}{revTwo}
\colorlet{revFourEdge}{revFour}
\colorlet{revEightEdge}{revEight}
\begin{figure}[!b]
\centering
\begin{minipage}{0.95\linewidth}
\begin{minipage}[t]{0.475\linewidth}
\vspace{0pt}
\centering
{\footnotesize\bfseries (a) Reverse branch attention mask $\Mb^{H,\mathrm r}$}.\par\smallskip
\resizebox{\linewidth}{!}{%
\begin{tikzpicture}[x=0.31cm,y=-0.31cm,font=\scriptsize,>=Stealth]
 \foreach \t in {0,...,14}{
  \foreach \s in {0,...,14}{
   \ifnum\s>\t\relax\else
    \pgfmathtruncatemacro{\lev}{floor(ln(\s+1)/ln(2)+0.0001)}
    \ifcase\lev\def\cellcolor{revOne}\or\def\cellcolor{revTwo}\or\def\cellcolor{revFour}\else\def\cellcolor{revEight}\fi
    \fill[\cellcolor,fill opacity=0.12] (\s,\t) rectangle (\s+1,\t+1);
   \fi
  }
 }
 \draw[step=1,black!10,line width=0.18pt] (0,0) grid (15,15);
 \draw[black!60,line width=0.5pt] (0,0) rectangle (15,15);
 \foreach \i in {0,...,14}{
  \node at (\i+0.5,-0.65) {\i};
  \node[anchor=east] at (-0.45,\i+0.5) {\i};
 }
 \node at (7.5,-2) {source position $s$};
 \node[rotate=90] at (-2.5,7) {query position $t$};
 \node[text=black!55,align=center] at (10.5,3.1) {future sources\\$s>t$: zero};
 \foreach \x in {1,3,7}{\draw[black!18,line width=0.25pt] (\x,0)--(\x,15);}
 \draw[revOneEdge,dash pattern=on 1.4pt off 1pt,line width=0.6pt]
  (0.05,0.05) rectangle (0.95,0.95);
 \draw[revOneEdge,line width=0.6pt] (0.05,1.05) rectangle (0.95,14.95);
 \draw[revTwoEdge,dash pattern=on 1.4pt off 1pt,line width=0.6pt]
  (1.05,1.05)--(1.95,1.05)--(1.95,2.05)--(2.95,2.05)--(2.95,2.95)--(1.05,2.95)--cycle;
 \draw[revTwoEdge,line width=0.6pt] (1.05,3.05) rectangle (2.95,14.95);
 \draw[revFourEdge,dash pattern=on 1.4pt off 1pt,line width=0.85pt]
  (3.05,3.05)--(3.95,3.05)--(3.95,4.05)--(4.95,4.05)--(4.95,5.05)--(5.95,5.05)--(5.95,6.05)--(6.95,6.05)--(6.95,6.95)--(3.05,6.95)--cycle;
 \draw[revFourEdge,line width=0.9pt] (3.05,7.05) rectangle (6.95,14.95);
 \draw[revEightEdge,dash pattern=on 1.4pt off 1pt,line width=0.6pt]
  (7.05,7.05)--(7.95,7.05)--(7.95,8.05)--(8.95,8.05)--(8.95,9.05)--(9.95,9.05)--(9.95,10.05)--(10.95,10.05)--(10.95,11.05)--(11.95,11.05)--(11.95,12.05)--(12.95,12.05)--(12.95,13.05)--(13.95,13.05)--(13.95,14.05)--(14.95,14.05)--(14.95,14.95)--(7.05,14.95)--cycle;
 \node[anchor=west,font=\scriptsize\bfseries] at (8.3,6.0) {(b)};
 \draw[->,black!70,line width=0.45pt] (8.1,6.0)--(6.6,6.0);
 \node[anchor=west,font=\scriptsize\bfseries] at (8.3,11.0) {(c)};
 \draw[->,black!70,line width=0.45pt] (8.1,11.0)--(6.6,11.0);
 \foreach \x/\y/\col/\ell/\range in {
  0/16.7/revOne/0/0,
  8/16.7/revTwo/1/{1,2},
  0/18.3/revFour/2/{3,\ldots,6},
  8/18.3/revEight/3/{7,\ldots,14}}{
  \filldraw[fill=\col,fill opacity=0.12,draw=\col,draw opacity=1,line width=0.6pt]
   (\x,\y-0.35) rectangle (\x+0.8,\y+0.35);
  \node[anchor=west] at (\x+1.1,\y) {$\cC_{\ell}=\{\range\}$};
 }
\end{tikzpicture}%
}
\end{minipage}\hfill
\begin{minipage}[t]{0.49\linewidth}
\vspace{0pt}
\footnotesize
\setlength{\arraycolsep}{2.5pt}
\renewcommand{\arraystretch}{1.12}
\setlength{\abovedisplayskip}{4pt}
\setlength{\belowdisplayskip}{4pt}
\textbf{(b) Within-bucket triangle for $\cC_2$}. Attention within this triangle is computed chunkwise (Algorithm~\ref{alg:reverse-chunk-attention}, lines~\ref{algline:chunk_loop_begin}--\ref{algline:chunk_loop_end}).
\vspace{2.5mm}
\[
 \Mb^{H,\mathrm r}_{3:7,\,3:7}
 =\begin{bmatrix}
 \mu_3^{(2)}&0&0&0\\
 \mu_4^{(2)}&\mu_4^{(2)}&0&0\\
 \mu_5^{(2)}&\mu_5^{(2)}&\mu_5^{(2)}&0\\
 \mu_6^{(2)}&\mu_6^{(2)}&\mu_6^{(2)}&\mu_6^{(2)}
 \end{bmatrix} = \operatorname{diag}\!\bigl(\bmu_{3:7}^{(2)}\bigr)
  \Mb_4^{\mathrm{causal}}
\]

\vspace{10mm}

\textbf{(c) Completed-bucket rectangle for $\cC_2$}. Later queries reuse the completed bucket summary (Algorithm~\ref{alg:reverse-chunk-attention}, lines~\ref{algline:after_bucket_begin}--\ref{algline:after_bucket_end}).
\vspace{2.5mm}
\[
 \Mb^{H,\mathrm r}_{7:15,\,3:7}
 =\begin{bmatrix}
 \mu_7^{(2)}&\mu_7^{(2)}&\mu_7^{(2)}&\mu_7^{(2)}\\
 \mu_8^{(2)}&\mu_8^{(2)}&\mu_8^{(2)}&\mu_8^{(2)}\\
 \vdots&\vdots&\vdots&\vdots\\
 \mu_{14}^{(2)}&\mu_{14}^{(2)}&\mu_{14}^{(2)}&\mu_{14}^{(2)}
 \end{bmatrix} = \bmu_{7:15}^{(2)}\oneb_4^\top.
\]
\end{minipage}
\end{minipage}
\caption{\textbf{Reverse branch attention mask structure and parallel computation.}
\textbf{(a)} Rows are queries and columns are source tokens for a 15-token sequence; distinct outline hues identify stationary source buckets ($\cC_0, \cC_1, \cC_2, \cC_3$), with faint matching cell fills. The stationary buckets match the reverse hierarchy in Figure~\ref{fig:bilateral-context-resolution}(b).
Dashed outlines mark within-bucket causal triangles; solid outlines mark completed-bucket rectangles. 
$\cC_3$ has no rectangle because no query follows its completion.
\textbf{(b,c)} The regions for $\cC_2$ are expanded into symbolic entries and compact forms. $\Mb_4^{\mathrm{causal}}$ is the $4\times4$ causal mask with ones on and below the diagonal and zeros elsewhere. Within each row, visible entries in a bucket share $\mu_t^{(\ell)}$, which may vary between queries. The triangle enforces causality while the bucket fills; the completed mask rectangle is an outer product, enabling reuse of its source summary in the attention computation. Ranges use half-open indexing, with $\bmu_{a:b}^{(2)}=(\mu_a^{(2)},\ldots,\mu_{b-1}^{(2)})^\top$. 
}
\label{fig:reverse-mask-structure}
\end{figure}
\endgroup

Algorithm~\ref{alg:reverse-chunk-attention} computes $\Ob^{\mathrm r}$ in Eq.~\eqref{eq:reverse-matrix} using chunks of at most $C$ tokens. It presents a sequential carry for clarity; an equivalent parallel evaluation is described afterwards. Subscripts select rows using half-open intervals. Let $\bm\mu^{(\ell)}=(\mu_0^{(\ell)},\ldots,\mu_{T-1}^{(\ell)})^\top$. The inclusive causal mask for a chunk of $c$ tokens satisfies $[\Mb_c^{\mathrm{causal}}]_{ij}=\mathbf1\{j\le i\}$, for $i,j=0,\ldots,c-1$. Diagonal multiplication denotes row-wise scaling; no diagonal matrix is materialized.

\begin{algorithm}[t]
\small
\DontPrintSemicolon
\SetCommentSty{textit}
\SetAlgoSkip{smallskip}
\caption{Reverse branch attention}
\label{alg:reverse-chunk-attention}
\KwIn{$\Qb,\Kb\in\RR^{T\times d_k}$; $\Vb\in\RR^{T\times d_v}$; write gates $\bm\eta\in\RR^T$; level weights $\{\bm\mu^{(\ell)}\}_{\ell=0}^{L_{\mathrm r}-1}$; chunk size $C\ge1$.}
\KwOut{Raw reverse attention output $\Ob^{\mathrm r}\in\RR^{T\times d_v}$.}
$\widetilde\Vb\gets\operatorname{diag}(\bm\eta)\Vb$\ \label{algline:value_gating}\\
$\Ob^{\mathrm r}\gets\zerob_{T\times d_v}$\;
\For{$\ell=0,\ldots,L_{\mathrm r}-1$}{
  $a\gets2^\ell-1$, $e\gets\min\{T,2^{\ell+1}-1\}$\;
  $\Rb\gets\zerob_{d_v\times d_k}$\;
  \vspace{0.1cm}
  \tcp{Within bucket: causal reads while the bucket fills; dashed triangles in Figure~\ref{fig:reverse-mask-structure}.}
  \ForEach{consecutive chunk $I$ in a partition of $[a,e)$, with $c=|I|\le C$}{ \label{algline:chunk_loop_begin}
    $\Qb_c,\Kb_c,\widetilde\Vb_c,\bm\mu_c\gets\Qb_I,\Kb_I,\widetilde\Vb_I,\bm\mu_I^{(\ell)}$\;
    $\Yb^{\mathrm{inter}}\gets\Qb_c\Rb^\top$\;
    $\Yb^{\mathrm{intra}}\gets[(\Qb_c\Kb_c^\top)\odot\Mb_c^{\mathrm{causal}}]\widetilde\Vb_c$\;
    $\Ob_I^{\mathrm r}\gets\Ob_I^{\mathrm r}+\operatorname{diag}(\bm\mu_c)(\Yb^{\mathrm{inter}}+\Yb^{\mathrm{intra}})$\;
    $\Rb\gets\Rb+\widetilde\Vb_c^\top\Kb_c$\; \label{algline:chunk_loop_end}
  }
  \vspace{0.1cm}
  \tcp{After bucket: later queries reuse the completed summary; solid rectangles in Figure~\ref{fig:reverse-mask-structure}.}
  \If{$e<T$}{ \label{algline:after_bucket_begin}
    $\Ob_{e:T}^{\mathrm r}\gets\Ob_{e:T}^{\mathrm r}+\operatorname{diag}(\bm\mu_{e:T}^{(\ell)})\Qb_{e:T}\Rb^\top$\; \label{algline:after_bucket_end}
  }
}
\KwRet{$\Ob^{\mathrm r}$}\;
\end{algorithm}

For a chunk of $c$ tokens, $\Qb_c,\Kb_c\in\RR^{c\times d_k}$, $\widetilde\Vb_c,\Yb^{\mathrm{inter}},\Yb^{\mathrm{intra}}\in\RR^{c\times d_v}$, and $\bm\mu_c\in\RR^c$. Before each chunk, $\Rb\in\RR^{d_v\times d_k}$ contains the gated writes from preceding chunks of the same bucket. The incoming-state read $\Yb^{\mathrm{inter}}$ and inclusive local read $\Yb^{\mathrm{intra}}$ cover disjoint parts of the visible bucket prefix. Reading $\Rb$ before adding the current chunk summary avoids double-counting tokens and including future tokens from that chunk. After the chunk loop, $\Rb$ summarizes the bucket's observed tokens and is reused by all later queries. The final bucket is clipped to $T$, so a partially filled final bucket has no later-query reads.

\paragraph{Computational cost.}
The buckets partition the $T$ source tokens. If their chunks have lengths $c_j\le C$, then $\sum_j c_j=T$ and $\sum_j c_j^2\le CT$. At fixed feature dimensions and $C$, value gating, within-bucket attention, state reads, and summary updates therefore take $O(T)$ total work. Each query reads at most $L_{\mathrm r}$ completed summaries in the after-bucket steps, giving at most $TL_{\mathrm r}$ such reads. Total arithmetic work is $O(TL_{\mathrm r})=O(T\log T)$, without constructing a global $T\times T$ attention matrix. Appendix~\ref{app:reverse-computation} provides dimensional work bounds and analyzes padding and training workspace; Section~\ref{sec:bilateral-cost} gives the bilateral cost comparison.

\paragraph{Parallel execution.}
Within each chunk, the incoming-state reads and masked local attention are evaluated by matrix multiplications that process all query positions in parallel. The sequential chunk loop in Algorithm~\ref{alg:reverse-chunk-attention} (line~\ref{algline:chunk_loop_begin}) can be replaced by an exclusive additive prefix scan within each bucket. We first compute the chunk summaries $\widetilde\Vb_c^\top\Kb_c$ independently. The scan then supplies each chunk with the sum of preceding chunk summaries, excluding its own. Once the incoming states are available, these operations can also be batched across chunks. The bucket total supports batched reads by all later queries. Ordinary addition suffices because reverse states have no temporal decay. Different levels maintain separate states, so their contributions can also be computed in parallel. A reduction across levels combines their overlapping output contributions before branch normalization. At fixed feature dimensions and chunk size, this reorganization preserves $O(T\log T)$ total arithmetic work. The full parallel formulation is provided in Appendix~\ref{app:reverse-computation}.

The final bucket summaries also supply the reverse states for recurrent continuation, including the summary of a partially filled final bucket. The next subsection combines $\Ob^{\mathrm r}$ with the inherited forward read $\Ob^{\mathrm f}$.

\subsection{Bilateral formulation}
\label{sec:bilateral-formulation}
\label{sec:bilateral-decoding}

The forward and reverse branches provide complementary reads of the same causal prefix. We combine their completed outputs through branch normalization and input-dependent weighting. All operations below are applied independently to each attention head; we suppress the head index, while retaining head-specific parameters and branch mixing coefficients.

Let $\bo_t^{\mathrm f},\bo_t^{\mathrm r}\in\RR^{d_v}$ denote the two raw reads. To reduce the influence of differences in their scales, we apply affine-free RMS normalization,
\begin{equation}
\hat{\bo}^f_t = \frac{\bo^f_t}{\sqrt{\lVert\bo^f_t\rVert_2^2/d_v+\varepsilon}},
\qquad \hat{\bo}^r_t = \frac{\bo^r_t}{\sqrt{\lVert\bo^r_t\rVert_2^2/d_v+\varepsilon}},
\qquad \varepsilon>0.
\label{eq:bilateral-branch-normalization}
\end{equation}
Normalization acts across the value channels of each token's read. It has no learned gain or bias and does not modify the stored memory matrices.

Let $\bu_t\in\RR^{d_{\mathrm{model}}}$ be the pre-normalized input to the layer. A learned affine branch router produces two logits and branch mixing coefficients using softmax:
\begin{equation}
\bg_t=\Wb_g\bu_t+\bb_g,
\qquad
(\rho_t^{\mathrm f},\rho_t^{\mathrm r})
 =\operatorname{softmax}_2(\bg_t),
\label{eq:bilateral-router}
\end{equation}
where $\Wb_g\in\RR^{2\times d_{\mathrm{model}}}$ and $\bb_g\in\RR^2$ denote the parameters for the head under consideration. The full layer produces two logits per head. The branch router depends on the current layer input, rather than directly on the branch reads. The branch mixing coefficients are positive and satisfy $\rho_t^{\mathrm f}+\rho_t^{\mathrm r}=1$. We initialize the branch router weights and biases to zero, giving equal branch mixing coefficients at initialization.

Using the hat notation of Eq.~\eqref{eq:bilateral-branch-normalization}, the bilateral attention read is:
\begin{equation}
\bo_t
 =\rho_t^{\mathrm f}\hat{\bo}_t^{\mathrm f}
 +\rho_t^{\mathrm r}\hat{\bo}_t^{\mathrm r}.
\label{eq:bilateral-read}
\end{equation}
The level weights $\lambda_t^{(\ell)}$ and $\mu_t^{(\ell)}$ select among summaries within each branch, whereas the branch mixing coefficients in Eq.~\eqref{eq:bilateral-router} weight the two normalized branch outputs. The reverse write gate $\eta_s$ instead controls the source contribution before storage. The softmax is over branches, rather than source positions; the query--key interactions in the raw reads remain unnormalized and may be signed.

For a sequence of length $T$, let $\bm{\rho}^a=(\rho_0^a,\ldots,\rho_{T-1}^a)^\top$ for $a\in\{\mathrm f,\mathrm r\}$. The same combination can be written as
\begin{equation}
\Ob
 =\operatorname{diag}(\bm{\rho}^{\mathrm f})\ \hat{\Ob}^{\mathrm f}
 +\operatorname{diag}(\bm{\rho}^{\mathrm r})\ \hat{\Ob}^{\mathrm r}
\label{eq:bilateral-matrix-combination}
\end{equation}
where hats denote independent row-wise normalization as in Eq.~\eqref{eq:bilateral-branch-normalization}: row $t$ of $\hat{\Ob}^a$ is $(\hat{\bo}_t^a)^\top$ for $a \in \{\mathrm f, \mathrm r\}$, and row $t$ of $\Ob$ is $(\bo_t)^\top$. The diagonal matrices express row-wise weighting and need not be materialized. 

During autoregressive decoding, we compute the shared current features, level weights, reverse write gate, and branch weight. We obtain the forward read using the inherited log-linear recurrence of \citet[Section~3.2]{guo2026loglinear}. We update and read the reverse states as described in Section~\ref{sec:reverse-recurrent}. We then apply Eq.~\eqref{eq:bilateral-read} to the two completed reads. During training, the inherited forward algorithm and Section~\ref{sec:reverse-parallel} provide the branch outputs, followed by the same row-wise normalization and combination in Eq.~\eqref{eq:bilateral-matrix-combination}. The two modes therefore implement the same bilateral attention computation.


\subsection{Computational cost and scope}
\label{sec:bilateral-cost}

The second hierarchy adds computation and recurrent storage while preserving the sequence-length scaling of log-linear attention. For a length-$T$ sequence, let
\[
L_{\mathrm f}=\lceil\log_2 T\rceil+1,
\qquad
L_{\mathrm r}=\lfloor\log_2 T\rfloor+1.
\]
Here $L_{\mathrm f}$ levels suffice for the forward hierarchy, consistent with Section~\ref{sec:background-loglinear}; some forward levels may be empty. The reverse hierarchy has $L_{\mathrm r}$ nonempty levels, one for each bucket defined in Section~\ref{sec:reverse-branch-attention}. Both counts are $O(\log T)$. Table~\ref{tab:bilateral-cost} separates the branch costs and their sum, treating feature dimensions and chunk size as fixed.

\begin{table}[tbp]
\centering
\small
\setlength{\tabcolsep}{5pt}
\caption{Attention-core work and matrix-state storage per head for a length-$T$ sequence, at fixed feature dimensions and chunk size. $L_{\mathrm f}$ levels suffice for the forward hierarchy, while $L_{\mathrm r}$ counts nonempty reverse levels; both grow logarithmically with $T$. Storage counts one matrix per level, including empty forward levels, and excludes shared feature-generation history, parameters, and training activations.}
\label{tab:bilateral-cost}
\begin{tabular}{@{}lccc@{}}
\toprule
 & Forward branch & Reverse branch & Bilateral \\
\midrule
Training/prefill work
 & $O(TL_{\mathrm f})$ & $O(TL_{\mathrm r})$
 & \shortstack{$O(T(L_{\mathrm f}+L_{\mathrm r}))=O(T\log T)$} \\[3pt]
Decode work per token
 & $O(L_{\mathrm f})$ & $O(L_{\mathrm r})$
 & \shortstack{$O(L_{\mathrm f}+L_{\mathrm r})=O(\log T)$} \\[3pt]
State matrices per head
 & $L_{\mathrm f}$ & $L_{\mathrm r}$
 & $L_{\mathrm f}+L_{\mathrm r}=O(\log T)$ \\[3pt]
Matrix-state elements per head
 & $L_{\mathrm f}d_vd_k$ & $L_{\mathrm r}d_vd_k$
 & \shortstack{$(L_{\mathrm f}+L_{\mathrm r})d_vd_k=O(\log T)$} \\
\bottomrule
\end{tabular}
\end{table}

The inherited forward algorithm and the reverse procedure in Section~\ref{sec:reverse-parallel} give $O(T(L_{\mathrm f}+L_{\mathrm r}))=O(T\log T)$ training/prefill work. During decoding, the reverse branch writes only one state, but reading its visible levels takes $O(L_{\mathrm r})$ work. Together with the forward update and read, bilateral decoding takes $O(L_{\mathrm f}+L_{\mathrm r})=O(\log T)$ work per token. Write gating, branch normalization, and fusion add constant work per token at fixed dimensions, without adding recurrent matrix-states. Past write gates are absorbed into the stored contributions and need no separate attention cache.

The forward and reverse hierarchies maintain $L_{\mathrm f}$ and $L_{\mathrm r}$ state matrices per head, respectively, giving $L_{\mathrm f}+L_{\mathrm r}$ matrices in total. Their storage is therefore $(L_{\mathrm f}+L_{\mathrm r})d_vd_k$ elements per head. Shared features avoid duplicating feature computation but do not remove the second matrix-state family.


\subsection{Parameterization}
\label{sec:bilateral-parameterization}

We use separate learned linear projections of the pre-normalized layer input $\bu_t$ to produce scalar level-weight signals. With head indices suppressed, these are
\begin{equation}
\begin{aligned}
d_t^{\lambda,(\ell)}&=(\ba_\lambda^{(\ell)})^\top\bu_t,\\
d_t^{\mu,(\ell)}&=(\ba_\mu^{(\ell)})^\top\bu_t,
\end{aligned}
\label{eq:bilateral-level-signals}
\end{equation}
where $\ba_\lambda^{(\ell)},\ba_\mu^{(\ell)}\in\RR^{d_{\mathrm{model}}}$ are learned independently for each head and level. No activation is applied to these signals. The positive level weights are then parameterized as
\begin{equation}
\begin{aligned}
\lambda_t^{(\ell)}
 &=\operatorname{softplus}\!\left(
 b^{\lambda,(\ell)}+s^{\lambda,(\ell)}d_t^{\lambda,(\ell)}\right),\\
\mu_t^{(\ell)}
 &=\operatorname{softplus}\!\left(
 b^{\mu,(\ell)}+s^{\mu,(\ell)}d_t^{\mu,(\ell)}\right).
\end{aligned}
\label{eq:bilateral-level-parameterization}
\end{equation}
The biases ($b^{\lambda,(\ell)}, b^{\mu,(\ell)}$) and sensitivities ($s^{\lambda,(\ell)}, s^{\mu,(\ell)}$) are learned independently for each head and level. The sensitivities need not be positive. Neither level-weight family is normalized across levels. 

An inverse-softplus bias sets the initial positive level-weight target, with a small nonzero input sensitivity allowing token dependence from initialization. 

The reverse write gate uses a separate learned affine projection of the input:
\begin{equation}
\eta_t=2\sigma\!\left(\ba_\eta^\top\bu_t+b_\eta\right),
\label{eq:reverse-write-gate}
\end{equation}
where $\sigma$ is the logistic sigmoid. Each head has its own projection vector $\ba_\eta\in\RR^{d_{\mathrm{model}}}$ and scalar bias $b_\eta$, with the head index suppressed. We initialize $\ba_\eta=\zerob$ and $b_\eta=0$, giving $\eta_t=1$ at initialization. The gate lies in $(0,2)$, so it can suppress or amplify a source write. It depends on the source input and is not normalized across tokens or levels. Setting $\eta_t=1$ for all sources recovers the ungated reverse branch.

The branch router and its equal-weight initialization are given in Eq.~\eqref{eq:bilateral-router}. We retain the inherited forward temporal dynamics without introducing another decay parameterization here. 

\section{Experiments}
\label{sec:experiments}

We evaluate bilateral log-linear attention on synthetic associative recall and language-modeling benchmarks. Our goal is to assess whether complementary context hierarchies improve recall under increasing memory load and support retrieval in pretrained language models while retaining language-modeling quality. Our primary comparison is among Mamba-2, forward log-linear attention, and bilateral log-linear attention, focusing on accuracy performance. These non-Transformer models summarize past context through recurrent states. We include a Transformer as a reference. 

The three non-Transformer models are implemented in pure PyTorch. The Transformer uses PyTorch's scaled dot-product attention (SDPA) operator, which provides a FlashAttention backend. Experiments were performed on a system with eight NVIDIA RTX PRO 6000 Blackwell Server Edition GPUs; individual run allocations are specified in the experimental details.

\subsection{Synthetic Benchmark Results}
\label{sec:synthetic-results}

We begin with multi-query associative recall (MQAR), a diagnostic of in-context retrieval \citep{arora2024zoology}. Each 512-token sequence contains $K$ distinct key--value pairs, presented once, followed by queries among random filler tokens. At each query, the model must predict the associated value. We use $K\in\{32,64,96,128\}$ to increase the number of associations stored within a fixed sequence length. Only query-value predictions contribute to the training loss and evaluation score.

\begin{figure}[tb]
\centering
\begin{minipage}[t]{0.49\linewidth}
  \centering
  \includegraphics[draft=false,width=\linewidth,height=0.65\linewidth]{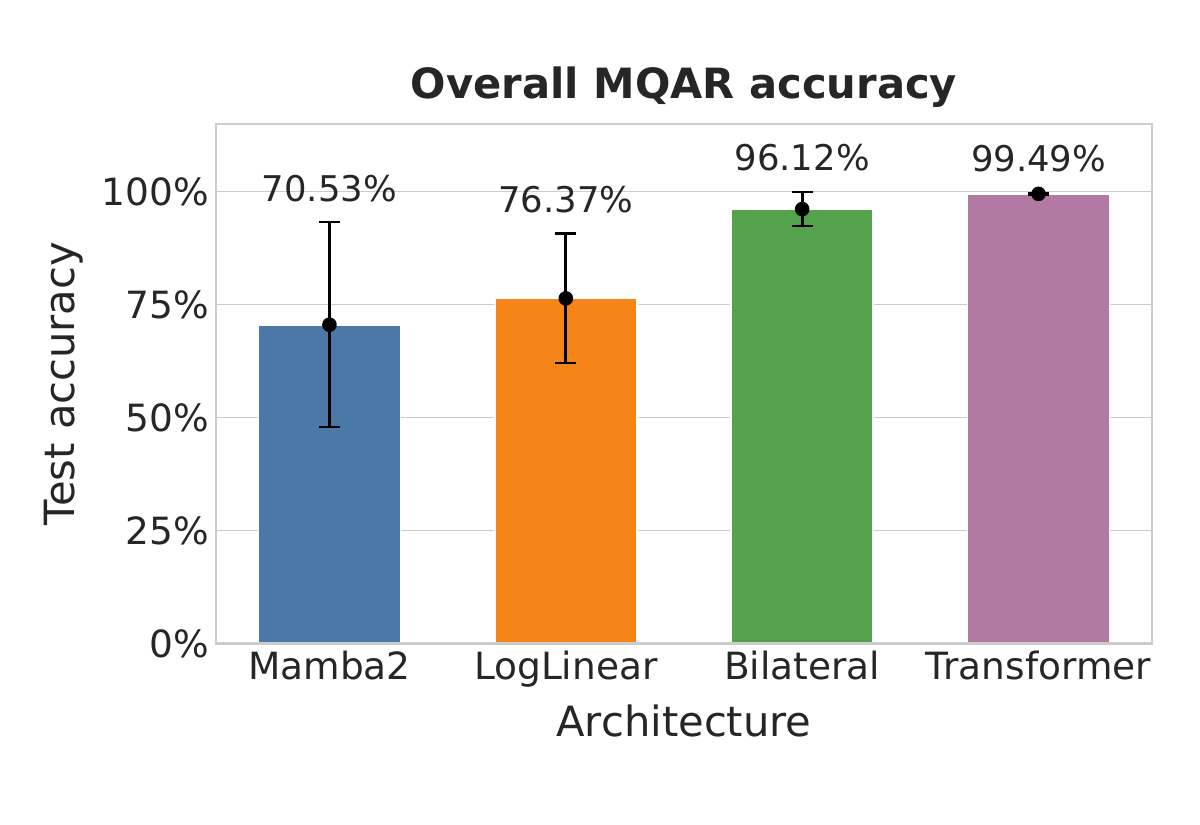}
  \par\smallskip
  {\small (a) Overall test accuracy.}
\end{minipage}\hfill
\begin{minipage}[t]{0.49\linewidth}
  \centering
  \includegraphics[draft=false,width=\linewidth,height=0.65\linewidth]{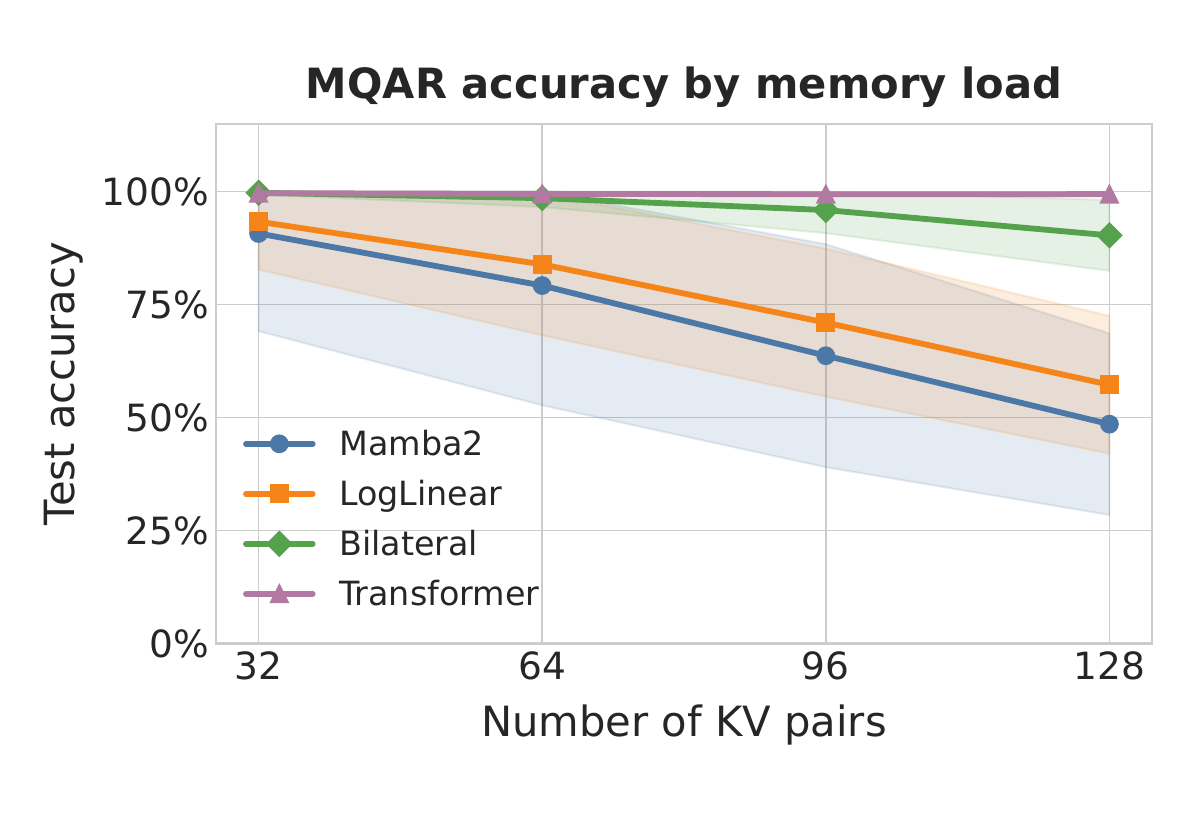}
  \par\smallskip
  {\small (b) Accuracy versus memory load.}
\end{minipage}
\caption{\textbf{MQAR test accuracy.} All four models have two layers and hidden size 64, with sequence length 512 and $K\in\{32,64,96,128\}$ key--value pairs. Bars and curves show means over ten final training seeds (10--19); error bars and shading show $\pm1$ sample standard deviation, with shaded bands clipped to $[0,100\%]$. All runs share the same fixed datasets, so the spread reflects variation across training seeds. Accuracy is averaged over queries within each example and then over examples; the overall score weights the four loads equally.}
\label{fig:mqar-results}
\end{figure}

All models have two layers and hidden size 64. The comparison matches depth and hidden width; parameter counts and block architectures differ. In particular, the Transformer includes a SwiGLU MLP, whereas the optional MLPs in the non-transformer models are disabled. Bilateral uses shared $\Qb, \Kb, \Vb$ features, branch RMS normalization, learned two-way softmax mixing, and the reverse write gate described in Section~\ref{sec:bilateral-loglinear}. Exact configurations and parameter counts are given in Appendix~\ref{app:mqar}.

For each memory load, we generate 100,000 training, 1000 validation, and 1000 test examples. All models and training seeds share the same fixed datasets. We select one learning rate per architecture using validation results over five training seeds, then train final models with ten disjoint training seeds for at most 50 epochs, stopping when validation accuracy exceeds 99\%. Each final run is evaluated once on the test set after restoring its best-validation checkpoint. Accuracy is averaged over queries within each example and then over examples; the overall score weights the four memory loads equally. Appendix~\ref{app:mqar} gives the full data, tuning, and evaluation protocol.

Figure~\ref{fig:mqar-results} shows that bilateral achieves $96.12\%$ mean overall test accuracy, with a sample standard deviation of $3.79$ percentage points across ten training seeds. Its mean exceeds forward log-linear attention by $19.75$ percentage points and Mamba-2 by $25.59$ points. Transformer has the highest overall mean, $99.49\%$, leaving a $3.37$-point gap to bilateral.

The advantage over the two state-space baselines grows with memory load. At $K=128$, bilateral retains $90.30\%$ mean accuracy, compared with $57.25\%$ for forward log-linear and $48.53\%$ for Mamba-2; Transformer reaches $99.42\%$. Bilateral also exhibits less variation across training seeds than those two baselines on the fixed datasets. The results are consistent with the motivation for complementary context resolution. Complete per-load results and overall scores for each training seed are reported in Appendix~\ref{app:mqar}.

\subsection{Language Modeling Results}
\label{sec:language-modeling-results}

\paragraph{Pretraining and evaluation setup.}
We pretrain on approximately 10B tokens from DHSA Long Data Collections\footnote{Dataset identifier: \texttt{sxiong/DHSA\_Long-Data-Collections}.}, using a training block length of 6144. All selected models use hidden width 1536, the same tokenizer, and training seed 1234. We compare Mamba-2, forward log-linear attention, bilateral log-linear attention, and Transformer in two settings: a \emph{layer-matched} comparison with 12 layers per model, and an \emph{approximately parameter-matched} comparison with 362.23M--370.07M learnable parameters. Table~\ref{tab:language-modeling-results} reports model sizes and aggregate results. The bilateral model uses shared features ($\Qb, \Kb, \Vb$), normalized softmax branch fusion, and the reverse write gate described in Section~\ref{sec:bilateral-loglinear}. We evaluate final checkpoints; the same bilateral checkpoint appears in both comparisons. Appendix~\ref{app:language-modeling} provides the tokenizer identifier, pretraining data and optimization details, exact parameter counts, evaluation settings, and task-level results.

\paragraph{Standard language modeling and commonsense reasoning.}
We report perplexity on WikiText-2~\citep{merity2016pointer} and LAMBADA~\citep{paperno2016lambada}, together with zero-shot scores from the LM Evaluation Harness~\citep{gao2021harness}. Following the metric selection of \citet{guo2026loglinear}, the LM-Eval average is the unweighted mean of six scores: LAMBADA accuracy, PIQA accuracy, HellaSwag length-normalized accuracy, WinoGrande accuracy, ARC-Easy accuracy, and ARC-Challenge length-normalized accuracy. Perplexities are excluded from this average. 
WikiText-2 rolling perplexity is evaluated with the same 6144-token context for all models, matching the training block length.

Bilateral achieves the lowest reported WikiText-2 and LAMBADA perplexities in both comparisons, at 30.21 and 33.20, respectively, under these evaluation settings. Its six-task LM-Eval average is 41.51\%, compared with 42.23\% for layer-matched log-linear attention and 41.26\% for approximately parameter-matched log-linear attention. The corresponding parameter-matched Mamba-2 and Transformer scores are 41.59\% and 41.80\%. Thus, the perplexity results are favorable, while the selected reasoning-task scores show mixed changes rather than uniform gains.

\begin{table}[t]
\centering
\caption{Language-modeling results under layer matching and approximate parameter matching. Parentheses give depth; parameter counts are in millions. PPL is perplexity ($\downarrow$); all other scores are percentages ($\uparrow$). LM-Eval averages six selected accuracy scores; RULER-NIAH averages six selected tasks separately at each length; retrieval averages eight tasks. Bold denotes the best result among Mamba-2, log-linear, and bilateral within each panel. The bilateral rows use the same checkpoint. 
WikiText-2 rolling PPL uses a 6144-token context for all models.
}
\label{tab:language-modeling-results}
\small
\setlength{\tabcolsep}{3pt}
\begin{tabular*}{\linewidth}{@{\extracolsep{\fill}}lrrrrrrrr@{}}
\toprule
& & \multicolumn{2}{c}{Perplexity $\downarrow$} & \multicolumn{1}{c}{LM-Eval} & \multicolumn{3}{c}{RULER-NIAH $\uparrow$} & Retrieval \\
\cmidrule(lr){3-4}\cmidrule(lr){6-8}
Model (layers) & Parameters (M) & WikiText-2 & LAMBADA & avg. $\uparrow$ & 2048 & 4096 & 6144 & avg. $\uparrow$ \\
\midrule
\multicolumn{9}{l}{\emph{Layer-matched (12 layers)}} \\
\addlinespace[3pt]
Mamba-2 (12) & 338.1 & 31.37 & 39.66 & 40.92 & 31.37 & 27.23 & 12.79 & 18.76 \\
Log-linear (12) & 350.5 & 30.23 & 33.78 & \textbf{42.23} & 43.65 & 40.52 & 27.04 & 21.46 \\
Bilateral (12) & 365.6 & \textbf{30.21} & \textbf{33.20} & 41.51 & \textbf{63.18} & \textbf{59.14} & \textbf{50.03} & \textbf{25.16} \\
\arrayrulecolor{black!40}\cmidrule(lr){1-9}\arrayrulecolor{black}
Transformer (12) & 275.7 & 32.53 & 40.75 & 40.81 & 67.42 & 61.28 & 59.77 & 29.76 \\
\midrule
\multicolumn{9}{l}{\emph{Approximately parameter-matched}} \\
\addlinespace[3pt]
Mamba-2 (13) & 362.2 & 30.93 & 33.97 & \textbf{41.59} & 43.62 & 30.15 & 18.44 & 20.35 \\
Log-linear (12) & 365.6 & 30.38 & 37.04 & 41.26 & 58.06 & 44.60 & 26.97 & 21.63 \\
Bilateral (12) & 365.6 & \textbf{30.21} & \textbf{33.20} & 41.51 & \textbf{63.18} & \textbf{59.14} & \textbf{50.03} & \textbf{25.16} \\
\arrayrulecolor{black!40}\cmidrule(lr){1-9}\arrayrulecolor{black}
Transformer (17) & 370.1 & 30.45 & 35.14 & 41.80 & 72.72 & 65.30 & 58.74 & 32.08 \\
\bottomrule
\end{tabular*}
\end{table}

\paragraph{Long-context retrieval.}
We evaluate six needle-in-a-haystack (NIAH) tasks from RULER~\citep{hsieh2024ruler} at context lengths 2048, 4096, and 6144, using the same 500 examples per task and length for every model. These lengths are within the training block length. Bilateral outperforms forward log-linear attention on all six selected NIAH tasks at each of the three context lengths, in both matching settings. Its advantage in the six-task average grows with context length. In the approximately parameter-matched setting, the gains are 5.12, 14.54, and 23.06 percentage points at the three lengths, respectively. At 6144 tokens, bilateral scores 50.03\%, compared with 26.97\% for log-linear attention and 18.44\% for Mamba-2. Transformer remains stronger on the aggregate, scoring 58.74\% at this length. Appendix~\ref{app:language-modeling} provides the complete task-level breakdown.

\paragraph{Downstream retrieval.}
We also evaluate SWDE, SQuAD completion, FDA, TriviaQA, DROP, and three Natural Questions (NQ) variants, denoted NQ-512, NQ-1024, and NQ-2048. The retrieval average is the unweighted mean of these eight tasks. All models use an 8192-token evaluation limit. 
Bilateral obtains a retrieval average of 25.16\%, exceeding layer-matched log-linear attention by 3.70 percentage points and approximately parameter-matched log-linear attention by 3.53 points. Transformer scores 29.76\% and 32.08\% in the respective comparisons. Bilateral outperforms log-linear attention on six of the eight retrieval tasks under layer matching and seven of eight under approximate parameter matching. Grouping the three NQ variants as one dataset, these correspond to four of six and five of six datasets, respectively; bilateral improves all three NQ variants in both comparisons. 

\section{Related Works}
\label{sec:related_works}

Linear attention summarizes past key--value associations in a recurrent matrix, enabling constant persistent storage with respect to sequence length~\citep{katharopoulos2020transformers}. RetNet~\citep{sun2023retentive}, H3~\citep{fu2023hungry}, RWKV~\citep{peng2023rwkv}, Gated Linear Attention~\citep{yang2024gated}, and Mamba-2~\citep{dao2024transformers} likewise admit recurrent representations whose size is independent of context length at fixed model dimensions. Gated DeltaNet combines adaptive forgetting with the delta rule for targeted memory updates~\citep{yang2025gateddelta}. Dynamic Linear Attention (DLA) adaptively determines state boundaries and merges adjacent summaries under a fixed maximum of $K$ states~\citep{wang2026dynamic}. Its partition adapts to content, while persistent memory remains bounded independently of context length at fixed $K$ and model dimensions. These approaches improve memory updates or allocation within bounded capacity. In earlier work, \citet{arora2024zoology} identify associative recall as an important source of the performance gap between the efficient architectures they study and attention, introducing multi-query associative recall to investigate it. \citet{arora2024based} further examine the tradeoff between recurrent state size and recall.

Full Transformer attention retains individually addressable key/value representations, with linear KV-cache growth and quadratic sequence-processing work at fixed feature dimensions~\citep{vaswani2017attention}. Sliding-window attention, used in Mistral~7B~\citep{jiang2023mistral}, limits direct attention to recent positions and supports an $O(W)$ rolling cache for window size $W$. Older entries are evicted, although their information can influence later representations through preceding layers. StreamingLLM additionally retains initial tokens as attention sinks to stabilize streaming generation~\citep{xiao2024streaming}. These methods bound storage by restricting which token-level entries remain accessible. Their retained initial and recent entries differ from a hierarchy of compressed summaries covering the causal prefix. More generally, sparsity in attention reads and reduction of persistent cache storage are separate properties.

BASED combines sliding-window and linear attention~\citep{arora2024based}, while Infini-attention integrates local attention with long-term compressive memory~\citep{munkhdalai2024infini}. At fixed window or segment sizes and feature dimensions, these hybrids retain bounded memory. Hierarchical predecessors include H-Transformer-1D's low-rank structure~\citep{zhu2021htransformer} and Fast Multipole Attention~\citep{kang2026multipole}, which preserves token-level attention locally while representing more distant groups through learned key and value summaries at progressively coarser resolutions. More directly, \citet{guo2026loglinear} introduce a Fenwick hierarchy of recurrent states with logarithmic storage and log-linear training work, providing finer grouping near the query and coarser grouping for earlier context. Adaptive Memory Decay enriches the parameterization of the input-dependent level weights using a two-layer MLP with softplus output, retaining the Fenwick partition~\citep{amin2026adaptive}. Prefix-scannable models~\citep{yau2026sequential} also support parallel evaluation and logarithmic online memory, using binary-tree aggregation that can accommodate non-associative operators. Their online algorithm merges completed blocks as the prefix grows. Our work extends log-linear attention with a complementary, origin-anchored partition whose source assignments remain fixed and whose completed summaries remain separately accessible without subsequent merging or temporal decay. Pairing this partition with the Fenwick hierarchy provides finer grouping near both the sequence origin and the current query. We derive equivalent recurrent and matrix forms and a chunkwise evaluation algorithm for the added branch, preserving the original asymptotic training work, decoding work, and persistent-state storage.

\section{Conclusion}
\label{sec:conclusion}

We introduced bilateral log-linear attention, which combines complementary causal hierarchies to support recall across the available context. The forward hierarchy provides finer resolution near the current query, while the origin-anchored hierarchy provides finer resolution near the beginning of the sequence. We derived equivalent recurrent and matrix formulations for the added branch and developed a chunkwise algorithm supporting parallel execution. At fixed feature dimensions and chunk size, the combined mechanism preserves log-linear training and prefill work, logarithmic decoding work per token, and logarithmic persistent state storage.

Our experiments demonstrate improved associative recall and retrieval relative to forward log-linear attention. With depth and hidden width matched, mean overall MQAR test accuracy increases from 76.37\% to 96.12\% across ten training seeds. Under approximate parameter matching, the mean over six selected RULER needle-in-a-haystack tasks increases from 26.97\% to 50.03\% at 6144 tokens, and bilateral attention outperforms the forward baseline on seven of eight downstream retrieval tasks. These findings support complementary context organization as a means of improving recall with compressed recurrent memory.

Our work highlights the broader principle of organizing memory to provide finer-resolution access near both ends of the causal prefix. This principle can be adapted to different state-update rules. In particular, a bilateral DeltaNet-style model could apply delta-rule updates to the states in both hierarchies and combine their outputs, preserving finer resolution near the current query and the sequence origin. The existing log-linear Gated DeltaNet construction provides a starting point for this extension~\citep{guo2026loglinear,yang2025gateddelta}. Deriving its recurrent and parallel computations and evaluating its performance are natural next steps beyond the present implementation. More broadly, this work highlights the allocation of context resolution as a design choice for improving recall in efficient sequence models.

\section*{Acknowledgments}
This research is sponsored by the FORESEE Initiative as part of the Laboratory Directed Research and Development Program of Oak Ridge National Laboratory, managed by UT-Battelle, LLC, for the U.S. Department of Energy under contract DE-AC05-00OR22725.

\bibliographystyle{plainnat}  
\bibliography{references}  

@inproceedings{
guo2026loglinear,
title={Log-Linear Attention},
author={Han Guo and Songlin Yang and Tarushii Goel and Eric P. Xing and Tri Dao and Yoon Kim},
booktitle={The Fourteenth International Conference on Learning Representations},
year={2026},
url={https://openreview.net/forum?id=mOJgZWkXKW}
}

@inproceedings{dao2024transformers,
author = {Dao, Tri and Gu, Albert},
title = {Transformers are {SSM}s: Generalized Models and Efficient Algorithms Through Structured State Space Duality},
year = {2024},
publisher = {PMLR},
booktitle = {Proceedings of the 41st International Conference on Machine Learning},
pages = {10041--10071},
volume = {235},
series = {Proceedings of Machine Learning Research},
url = {https://proceedings.mlr.press/v235/dao24a.html}
}

@inproceedings{arora2024zoology,
  title={Zoology: Measuring and improving recall in efficient language models},
  author={Arora, Simran and Eyuboglu, Sabri and Timalsina, Aman and Johnson, Isys and Poli, Michael and Zou, James Y and Rudra, Atri and R{\'e}, Christopher},
  booktitle={The Twelfth International Conference on Learning Representations},
  year={2024},
  url={https://arxiv.org/abs/2312.04927}
}

@inproceedings{katharopoulos2020transformers,
  title={Transformers are {RNN}s: Fast Autoregressive Transformers with Linear Attention},
  author={Katharopoulos, Angelos and Vyas, Apoorv and Pappas, Nikolaos and Fleuret, Fran{\c{c}}ois},
  booktitle={Proceedings of the 37th International Conference on Machine Learning},
  pages={5156--5165},
  volume={119},
  series={Proceedings of Machine Learning Research},
  year={2020},
  publisher={PMLR},
  url={https://proceedings.mlr.press/v119/katharopoulos20a.html}
}

@misc{gao2021harness,
  author = {Gao, Leo and Tow, Jonathan and Abbasi, Baber and Biderman, Stella and Black, Sid and DiPofi, Anthony and Foster, Charles and Golding, Laurence and Hsu, Jeffrey and Le Noac'h, Alain and Li, Haonan and McDonell, Kyle and Muennighoff, Niklas and Ociepa, Chris and Phang, Jason and Reynolds, Laria and Schoelkopf, Hailey and Skowron, Aviya and Sutawika, Lintang and Tang, Eric and Thite, Anish and Wang, Ben and Wang, Kevin and Zou, Andy},
  title = {A framework for few-shot language model evaluation},
  month = sep,
  year = {2021},
  publisher = {Zenodo},
  version = {v0.0.1},
  doi = {10.5281/zenodo.5371628},
  url = {https://doi.org/10.5281/zenodo.5371628}
}

@inproceedings{hsieh2024ruler,
  author = {Hsieh, Cheng-Ping and Sun, Simeng and Kriman, Samuel and Acharya, Shantanu and Rekesh, Dima and Jia, Fei and Zhang, Yang and Ginsburg, Boris},
  title = {{RULER}: What's the Real Context Size of Your Long-Context Language Models?},
  booktitle = {Conference on Language Modeling},
  year = {2024},
  url = {https://arxiv.org/abs/2404.06654}
}

@misc{merity2016pointer,
  author = {Merity, Stephen and Xiong, Caiming and Bradbury, James and Socher, Richard},
  title = {Pointer Sentinel Mixture Models},
  year = {2016},
  eprint = {1609.07843},
  archivePrefix = {arXiv},
  url = {https://arxiv.org/abs/1609.07843}
}

@inproceedings{paperno2016lambada,
  author = {Paperno, Denis and Kruszewski, Germ{\'a}n and Lazaridou, Angeliki and Pham, Ngoc Quan and Bernardi, Raffaella and Pezzelle, Sandro and Baroni, Marco and Boleda, Gemma and Fern{\'a}ndez, Raquel},
  title = {The {LAMBADA} dataset: Word prediction requiring a broad discourse context},
  booktitle = {Proceedings of the 54th Annual Meeting of the Association for Computational Linguistics (Volume 1: Long Papers)},
  year = {2016},
  pages = {1525--1534},
  doi = {10.18653/v1/P16-1144},
  url = {https://aclanthology.org/P16-1144/}
}

@inproceedings{vaswani2017attention,
  title = {Attention Is All You Need},
  author = {Vaswani, Ashish and Shazeer, Noam and Parmar, Niki and Uszkoreit, Jakob and Jones, Llion and Gomez, Aidan N. and Kaiser, {\L}ukasz and Polosukhin, Illia},
  booktitle = {Advances in Neural Information Processing Systems},
  volume = {30},
  year = {2017},
  url = {https://proceedings.neurips.cc/paper/2017/hash/3f5ee243547dee91fbd053c1c4a845aa-Abstract.html}
}

@misc{sun2023retentive,
  title = {Retentive Network: A Successor to Transformer for Large Language Models},
  author = {Sun, Yutao and Dong, Li and Huang, Shaohan and Ma, Shuming and Xia, Yuqing and Xue, Jilong and Wang, Jianyong and Wei, Furu},
  year = {2023},
  eprint = {2307.08621},
  archivePrefix = {arXiv},
  primaryClass = {cs.CL},
  url = {https://arxiv.org/abs/2307.08621}
}

@inproceedings{yang2024gated,
  title = {Gated Linear Attention Transformers with Hardware-Efficient Training},
  author = {Yang, Songlin and Wang, Bailin and Shen, Yikang and Panda, Rameswar and Kim, Yoon},
  booktitle = {Proceedings of the 41st International Conference on Machine Learning},
  pages = {56501--56523},
  year = {2024},
  volume = {235},
  series = {Proceedings of Machine Learning Research},
  publisher = {PMLR},
  url = {https://proceedings.mlr.press/v235/yang24ab.html}
}

@article{kwiatkowski2019natural,
  title = {{Natural Questions}: A Benchmark for Question Answering Research},
  author = {Kwiatkowski, Tom and Palomaki, Jennimaria and Redfield, Olivia and Collins, Michael and Parikh, Ankur and Alberti, Chris and Epstein, Danielle and Polosukhin, Illia and Devlin, Jacob and Lee, Kenton and Toutanova, Kristina and Jones, Llion and Kelcey, Matthew and Chang, Ming-Wei and Dai, Andrew M. and Uszkoreit, Jakob and Le, Quoc and Petrov, Slav},
  journal = {Transactions of the Association for Computational Linguistics},
  volume = {7},
  year = {2019},
  pages = {452--466},
  doi = {10.1162/tacl_a_00276},
  url = {https://aclanthology.org/Q19-1026/}
}

@inproceedings{fu2023hungry,
  title={Hungry Hungry Hippos: Towards Language Modeling with State Space Models},
  author={Fu, Daniel Y. and Dao, Tri and Saab, Khaled K. and Thomas, Armin W. and Rudra, Atri and R{\'e}, Christopher},
  booktitle={The Eleventh International Conference on Learning Representations},
  year={2023},
  url={https://arxiv.org/abs/2212.14052}
}

@inproceedings{peng2023rwkv,
  title={{RWKV}: Reinventing {RNN}s for the Transformer Era},
  author={Peng, Bo and Alcaide, Eric and Anthony, Quentin and Albalak, Alon and Arcadinho, Samuel and Biderman, Stella and Cao, Huanqi and Cheng, Xin and Chung, Michael and Derczynski, Leon and Du, Xingjian and Grella, Matteo and Gv, Kranthi and He, Xuzheng and Hou, Haowen and Kazienko, Przemyslaw and Kocon, Jan and Kong, Jiaming and Koptyra, Bart{\l}omiej and Lau, Hayden and Lin, Jiaju and Mantri, Krishna Sri Ipsit and Mom, Ferdinand and Saito, Atsushi and Song, Guangyu and Tang, Xiangru and Wind, Johan and Wo{\'z}niak, Stanis{\l}aw and Zhang, Zhenyuan and Zhou, Qinghua and Zhu, Jian and Zhu, Rui-Jie},
  booktitle={Findings of the Association for Computational Linguistics: EMNLP 2023},
  year={2023},
  pages={14048--14077},
  publisher={Association for Computational Linguistics},
  doi={10.18653/v1/2023.findings-emnlp.936},
  url={https://aclanthology.org/2023.findings-emnlp.936/}
}

@inproceedings{arora2024based,
  title={Simple linear attention language models balance the recall-throughput tradeoff},
  author={Arora, Simran and Eyuboglu, Sabri and Zhang, Michael and Timalsina, Aman and Alberti, Silas and Zou, James and Rudra, Atri and R{\'e}, Christopher},
  booktitle={Proceedings of the 41st International Conference on Machine Learning},
  pages={1763--1840},
  volume={235},
  series={Proceedings of Machine Learning Research},
  year={2024},
  publisher={PMLR},
  url={https://proceedings.mlr.press/v235/arora24a.html}
}

@inproceedings{yang2025gateddelta,
  title={Gated Delta Networks: Improving {Mamba2} with Delta Rule},
  author={Yang, Songlin and Kautz, Jan and Hatamizadeh, Ali},
  booktitle={The Thirteenth International Conference on Learning Representations},
  year={2025},
  url={https://arxiv.org/abs/2412.06464}
}

@misc{jiang2023mistral,
  title={{Mistral 7B}},
  author={Jiang, Albert Q. and Sablayrolles, Alexandre and Mensch, Arthur and Bamford, Chris and Chaplot, Devendra Singh and de las Casas, Diego and Bressand, Florian and Lengyel, Gianna and Lample, Guillaume and Saulnier, Lucile and Renard Lavaud, L{\'e}lio and Lachaux, Marie-Anne and Stock, Pierre and Le Scao, Teven and Lavril, Thibaut and Wang, Thomas and Lacroix, Timoth{\'e}e and El Sayed, William},
  year={2023},
  eprint={2310.06825},
  archivePrefix={arXiv},
  primaryClass={cs.CL},
  url={https://arxiv.org/abs/2310.06825}
}

@inproceedings{xiao2024streaming,
  title={Efficient Streaming Language Models with Attention Sinks},
  author={Xiao, Guangxuan and Tian, Yuandong and Chen, Beidi and Han, Song and Lewis, Mike},
  booktitle={The Twelfth International Conference on Learning Representations},
  year={2024},
  url={https://arxiv.org/abs/2309.17453}
}

@misc{munkhdalai2024infini,
  title={Leave No Context Behind: Efficient Infinite Context Transformers with {Infini-attention}},
  author={Munkhdalai, Tsendsuren and Faruqui, Manaal and Gopal, Siddharth},
  year={2024},
  eprint={2404.07143},
  archivePrefix={arXiv},
  primaryClass={cs.CL},
  url={https://arxiv.org/abs/2404.07143}
}

@inproceedings{zhu2021htransformer,
  title={{H-Transformer-1D}: Fast One-Dimensional Hierarchical Attention for Sequences},
  author={Zhu, Zhenhai and Soricut, Radu},
  booktitle={Proceedings of the 59th Annual Meeting of the Association for Computational Linguistics and the 11th International Joint Conference on Natural Language Processing (Volume 1: Long Papers)},
  pages={3801--3815},
  year={2021},
  publisher={Association for Computational Linguistics},
  doi={10.18653/v1/2021.acl-long.294},
  url={https://aclanthology.org/2021.acl-long.294/}
}

@article{kang2026multipole,
  title={Fast Multipole Attention: A Scalable Multilevel Attention Mechanism for Text and Images},
  author={Kang, Yanming and Tran, Giang and De Sterck, Hans},
  journal={Neural Networks},
  pages={109847},
  year={2026},
  note={In press},
  doi={10.1016/j.neunet.2026.109847},
  url={https://doi.org/10.1016/j.neunet.2026.109847}
}

@inproceedings{yau2026sequential,
  title={Sequential Parallel Duality in Prefix Scannable Models},
  author={Yau, Morris and Gupta, Sharut and Engelmayer, Valerie and Irie, Kazuki and Jegelka, Stefanie and Andreas, Jacob},
  booktitle={The Fourteenth International Conference on Learning Representations},
  year={2026},
  url={https://openreview.net/forum?id=tuLF84azND}
}

@inproceedings{wang2026dynamic,
  title={Dynamic Linear Attention},
  author={Wang, Xin and Shen, Hui and Zheng, Boyuan and Liu, Xueshen and Cho, Minkyoung and Wan, Zhongwei and Zhao, Zesen and Mao, Zhuoqing and Yan, Shen and Zhang, Mi},
  booktitle={Proceedings of the 43rd International Conference on Machine Learning},
  pages={129740--129752},
  volume={306},
  series={Proceedings of Machine Learning Research},
  year={2026},
  publisher={PMLR},
  url={https://proceedings.mlr.press/v306/wang26gr.html}
}

@misc{amin2026adaptive,
  title={Adaptive Memory Decay for Log-Linear Attention},
  author={Amin, Yaxita and Li, Helen Zichen and Zhang, Mengfan and Ayhan, Samet},
  year={2026},
  eprint={2605.06946},
  archivePrefix={arXiv},
  primaryClass={cs.LG},
  note={Preprint},
  url={https://arxiv.org/abs/2605.06946}
}

\clearpage
\appendix
\section{Reverse Branch Attention: Parallel Evaluation and Complexity}
\label{app:reverse-computation}

This appendix derives a scan-based evaluation of the reverse operator in Eq.~\eqref{eq:reverse-matrix}. The primary algorithm keeps a sequential loop over levels and evaluates chunks within each level in parallel, following the schedule of our implementation. We explain its relationship to Algorithm~\ref{alg:reverse-chunk-attention}, describe optional parallel evaluation across levels, and analyze arithmetic work and storage. The forward branch uses the algorithm of \citet{guo2026loglinear}.

\subsection{Notation and input contract}
\label{app:reverse-scan-notation}

We consider one sequence and one attention head, with $T\ge1$, $\Qb,\Kb\in\RR^{T\times d_k}$, and $\Vb\in\RR^{T\times d_v}$. Source gates are applied once:
\[
\widetilde\Vb=\operatorname{diag}(\bm\eta)\Vb.
\]
The mathematical algorithm below receives base values and gates explicitly; the implementation receives $\widetilde\Vb$ from its caller. The query-dependent weights $\mu_t^{(\ell)}$ are the same effective level weights as in Eq.~\eqref{eq:reverse-state-read}. They scale reads, not the source writes accumulated by the scan.

There are $L_{\mathrm r}=\lfloor\log_2T\rfloor+1$ nonempty reverse buckets. For level $\ell$, define the observed interval and its length as
\begin{equation}
a_\ell=2^\ell-1,\qquad
e_\ell=\min\{T,2^{\ell+1}-1\},\qquad
W_\ell=[a_\ell,e_\ell),\qquad m_\ell=e_\ell-a_\ell.
\label{eq:reverse-appendix-intervals}
\end{equation}
Intervals index integer token positions and are half-open. For a positive integer chunk size $C$, partition $W_\ell$ into $N_\ell=\lceil m_\ell/C\rceil$ consecutive chunks,
\begin{equation}
I_{\ell,j}=[a_\ell+jC,\,\min\{a_\ell+(j+1)C,e_\ell\}),
\qquad c_{\ell,j}=|I_{\ell,j}|,\qquad 0\le j<N_\ell.
\label{eq:reverse-appendix-chunks}
\end{equation}
We abbreviate row selections by $\Qb_{\ell,j}=\Qb_{I_{\ell,j}}$, and similarly for $\Kb$, $\widetilde\Vb$, and $\bm\mu^{(\ell)}$. The local mask is inclusive: $[\Mb_c^{\mathrm{causal}}]_{uv}=\mathbf1\{v\le u\}$. Diagonal matrices below denote row scaling and need not be materialized.

\subsection{Exclusive scans and batched reads}
\label{app:reverse-scan-formulation}

Each chunk independently forms a gated value--key summary. Its incoming state contains only preceding chunks of the same bucket:
\begin{equation}
\Ab_{\ell,j}=\widetilde\Vb_{\ell,j}^{\top}\Kb_{\ell,j},\qquad
\Rb_{\ell,j}=\sum_{i<j}\Ab_{\ell,i},\qquad
\Sbb_\ell=\sum_{j=0}^{N_\ell-1}\Ab_{\ell,j}.
\label{eq:reverse-appendix-scan}
\end{equation}
All three matrices have shape $d_v\times d_k$, and $\Rb_{\ell,0}=\zerob$. An exclusive additive prefix scan computes all $\Rb_{\ell,j}$. Equivalently, an inclusive scan gives $\Pb_{\ell,j}=\sum_{i\le j}\Ab_{\ell,i}$; shifting these prefixes by one position and prepending zero gives the incoming states, while $\Sbb_\ell=\Pb_{\ell,N_\ell-1}$. Our implementation instead forms $\Rb_{\ell,j}=\Pb_{\ell,j}-\Ab_{\ell,j}$, which is algebraically equivalent. Floating-point reassociation and subtraction can produce numerical differences; equivalence here is in real arithmetic, not bitwise equality. No decay factors or query-dependent level weights enter the scan.

Let $\Ob^{\mathrm r,(\ell)}$ denote the contribution of level $\ell$ to the raw reverse output. Its within-bucket rows are
\begin{equation}
\Ob_{I_{\ell,j}}^{\mathrm r,(\ell)}
=\operatorname{diag}(\bm\mu_{\ell,j})\left[
\underbrace{\Qb_{\ell,j}\Rb_{\ell,j}^{\top}}_{\text{preceding chunks}}
+\underbrace{\bigl((\Qb_{\ell,j}\Kb_{\ell,j}^{\top})\odot
\Mb_{c_{\ell,j}}^{\mathrm{causal}}\bigr)\widetilde\Vb_{\ell,j}}_{\text{inclusive local read}}
\right].
\label{eq:reverse-appendix-chunk-read}
\end{equation}
The local reads can be computed independently of the scan. Once incoming states are available, all preceding-chunk reads can also be batched. Matrix multiplications evaluate all query positions within each chunk without a sequential token loop.

For the later-query interval $E_\ell=[e_\ell,T)$, the completed summary gives
\begin{equation}
\Ob_{E_\ell}^{\mathrm r,(\ell)}
=\operatorname{diag}(\bm\mu_{E_\ell}^{(\ell)})\Qb_{E_\ell}\Sbb_\ell^{\top}.
\label{eq:reverse-appendix-later-read}
\end{equation}
These reads require the bucket total but not the within-bucket outputs. If $E_\ell$ is empty, this step is omitted. Rows before $a_\ell$ receive zero contribution from level $\ell$.

Equation~\eqref{eq:reverse-appendix-later-read} follows from the completed-bucket rectangle in Figure~\ref{fig:reverse-mask-structure}. For nonempty $E_\ell$,
\begin{equation}
\begin{aligned}
\Mb^{H,\mathrm r}_{E_\ell,W_\ell}
 &=\bm\mu_{E_\ell}^{(\ell)}\mathbf1_{W_\ell}^{\top},\\
\Ob_{E_\ell}^{\mathrm r,(\ell)}
 &=\operatorname{diag}(\bm\mu_{E_\ell}^{(\ell)})\Qb_{E_\ell}
 \bigl(\Kb_{W_\ell}^{\top}\operatorname{diag}(\bm\eta_{W_\ell})\Vb_{W_\ell}\bigr).
\end{aligned}
\label{eq:reverse-appendix-factorization}
\end{equation}
The parenthesized matrix is $\Sbb_\ell^\top$. The mask rectangle has rank one because its level weights are positive. The corresponding query--key-weighted block has rank at most $d_k$; multiplying by source gates preserves this upper bound. Thus all later queries reuse one summary instead of revisiting the bucket's individual source tokens.

\subsection{Primary algorithm: sequential levels, parallel chunks}
\label{app:reverse-scan-algorithm}

Algorithm~\ref{alg:reverse-scan-attention} replaces the sequential chunk carry in Algorithm~\ref{alg:reverse-chunk-attention} with an exclusive scan. Its outer loop remains sequential. Within each level, distinct chunks write disjoint output rows, and the later-query interval is disjoint from the bucket. Contributions from different levels can overlap, but the sequential outer loop makes their accumulation into one output buffer well defined.

\begin{algorithm}[htbp]
\small
\DontPrintSemicolon
\SetCommentSty{textit}
\SetAlgoSkip{smallskip}
\caption{Reverse attention with parallel chunks and sequential levels}
\label{alg:reverse-scan-attention}
\KwIn{$\Qb,\Kb\in\RR^{T\times d_k}$; $\Vb\in\RR^{T\times d_v}$; write gates $\bm\eta$; level weights $\{\bm\mu^{(\ell)}\}_{\ell=0}^{L_{\mathrm r}-1}$; chunk size $C\ge1$.}
\KwOut{Raw output $\Ob^{\mathrm r}$ and final bucket summaries $\{\Sbb_\ell\}_{\ell=0}^{L_{\mathrm r}-1}$.}
$\widetilde\Vb\gets\operatorname{diag}(\bm\eta)\Vb$\;
$\Ob^{\mathrm r}\gets\zerob_{T\times d_v}$\;
\For{$\ell=0,\ldots,L_{\mathrm r}-1$ \textnormal{(sequential)}}{
  Form $W_\ell$, $E_\ell$, and chunks $\{I_{\ell,j}\}_{j=0}^{N_\ell-1}$ using Eqs.~\eqref{eq:reverse-appendix-intervals}--\eqref{eq:reverse-appendix-chunks}\;
  \ForEach{$j=0,\ldots,N_\ell-1$ \textnormal{in parallel}}{
    $\Ab_{\ell,j}\gets\widetilde\Vb_{\ell,j}^{\top}\Kb_{\ell,j}$\;
    $\Yb_{\ell,j}^{\mathrm{intra}}\gets[(\Qb_{\ell,j}\Kb_{\ell,j}^{\top})\odot\Mb_{c_{\ell,j}}^{\mathrm{causal}}]\widetilde\Vb_{\ell,j}$\;
  }
  $(\Rb_{\ell,0},\ldots,\Rb_{\ell,N_\ell-1})\gets\operatorname{ExclusiveSum}(\Ab_{\ell,0},\ldots,\Ab_{\ell,N_\ell-1})$\;
  $\Sbb_\ell\gets\Rb_{\ell,N_\ell-1}+\Ab_{\ell,N_\ell-1}$\;
  \ForEach{$j=0,\ldots,N_\ell-1$ \textnormal{in parallel}}{
    $\Yb_{\ell,j}^{\mathrm{inter}}\gets\Qb_{\ell,j}\Rb_{\ell,j}^{\top}$\;
    $\Ob_{I_{\ell,j}}^{\mathrm r}\gets\Ob_{I_{\ell,j}}^{\mathrm r}+\operatorname{diag}(\bm\mu_{\ell,j})(\Yb_{\ell,j}^{\mathrm{inter}}+\Yb_{\ell,j}^{\mathrm{intra}})$\;
  }
  \If{$e_\ell<T$}{
    $\Ob_{E_\ell}^{\mathrm r}\gets\Ob_{E_\ell}^{\mathrm r}+\operatorname{diag}(\bm\mu_{E_\ell}^{(\ell)})\Qb_{E_\ell}\Sbb_\ell^\top$\;
  }
}
\KwRet{$\Ob^{\mathrm r},\{\Sbb_\ell\}_{\ell=0}^{L_{\mathrm r}-1}$}\;
\end{algorithm}

The exclusive scan supplies each chunk with the gated writes from preceding chunks of the same bucket, exactly as the sequential carry in Algorithm~\ref{alg:reverse-chunk-attention}. The inclusive local mask covers the current chunk's causal prefix. These disjoint contributions recover the visible bucket state, including the current token, without double counting or reading future tokens. Later queries read the completed bucket summary. Summing the level contributions therefore recovers the raw reverse output in Eq.~\eqref{eq:reverse-matrix}.

\paragraph{Optional parallel execution across levels.}
Each level depends only on shared input features and its own source summaries. Levels may therefore be evaluated concurrently by forming their contributions separately and reducing them:
\begin{equation}
\Ob^{\mathrm r}=\sum_{\ell=0}^{L_{\mathrm r}-1}\Ob^{\mathrm r,(\ell)}.
\label{eq:reverse-appendix-level-reduction}
\end{equation}
This transformation requires coordinated accumulation, such as a reduction over separate level outputs; simply making the outer loop parallel while retaining unsynchronized updates to $\Ob^{\mathrm r}$ would not suffice. The reduction precedes the branch normalization in Eq.~\eqref{eq:bilateral-branch-normalization}. Normalizing individual level contributions would change the operator. Materializing all level outputs is a simple reference schedule; tiled accumulation can reduce temporary storage with more involved scheduling. Batch and head dimensions offer additional parallelism. These schedules evaluate a sequence whose features are available, as in training or prefill; they do not generate unknown future tokens in parallel.

Our implementation uses sequential levels. Concurrent levels are a theoretical scheduling option, not a source of measured speedup claimed here. Exponentially increasing bucket sizes alone do not determine its benefit: small buckets still serve many later queries through Eq.~\eqref{eq:reverse-appendix-later-read}. Available hardware resources, batch and head parallelism, scheduling overhead, and temporary-output traffic determine whether concurrent levels improve runtime.

\subsection{Arithmetic work}
\label{app:reverse-scan-cost}

We count forward evaluation of the reverse attention core, excluding feature projections, generation of gates and level weights, and subsequent branch normalization and fusion. Let
\begin{equation}
N=\sum_\ell N_\ell,\qquad
U=\sum_\ell\sum_{j=0}^{N_\ell-1}c_{\ell,j}^2,\qquad
F=\sum_\ell(T-e_\ell).
\label{eq:reverse-appendix-work-counts}
\end{equation}
The buckets partition $T$ tokens, so $\sum_\ell m_\ell=T$, $N\le T$, and $U\le CT$. The count $F$ includes one completed-state read for every later query at each level. In particular,
\begin{equation}
F=(L_{\mathrm r}-1)(T+1)-2^{L_{\mathrm r}}+2
\le T(L_{\mathrm r}-1).
\label{eq:reverse-appendix-later-count}
\end{equation}
Here every level except the last has $e_\ell=2^{\ell+1}-1$, and the last has $e_\ell=T$.

\begin{table}[tbp]
\centering
\small
\caption{Arithmetic work per sequence and head for the unpadded scan algorithm. Scan bounds assume a work-efficient additive scan. Scaling and accumulation include both within-bucket and later-query rows.}
\label{tab:reverse-scan-work}
\begin{tabular}{@{}ll@{}}
\toprule
Operation & Arithmetic work \\
\midrule
Apply source gates & $O(Td_v)$ \\
Form chunk summaries & $O(Td_kd_v)$ \\
Scan summaries and obtain bucket totals & $O(Nd_kd_v)$ \\
Local causal query--key/value products & $O(U(d_k+d_v))$ \\
Read incoming chunk states & $O(Td_kd_v)$ \\
Read completed states at later queries & $O(Fd_kd_v)$ \\
Apply level weights and accumulate outputs & $O((T+F)d_v)$ \\
\bottomrule
\end{tabular}
\end{table}

Table~\ref{tab:reverse-scan-work} yields the dimensional bound
\begin{equation}
O\!\left(TC(d_k+d_v)+(T+F)d_kd_v\right).
\label{eq:reverse-appendix-dimensional-work}
\end{equation}
We have absorbed the scan, gating, and accumulation terms using $N\le T$ and $d_k,d_v\ge1$. For fixed dimensions and $C$, this is $O(TL_{\mathrm r})=O(T\log(1+T))$, or $O(T\log T)$ for $T\ge2$; the single-token case has constant work. The optional dense reduction across level outputs adds $O(TL_{\mathrm r}d_v)$ work and preserves this sequence-length scaling. No global attention matrix is needed. For $B$ sequences of the same length and $H$ heads of the same dimensions, the work bounds acquire a factor $BH$.

\subsection{Padding, storage, and recurrent continuation}
\label{app:reverse-scan-boundaries}

\paragraph{Padding and boundary cases.}
Algorithm~\ref{alg:reverse-scan-attention} uses actual chunk lengths. The implementation instead pads each bucket to $\widetilde m_\ell=C\lceil m_\ell/C\rceil$ and batches equal-size chunks. It appends zero rows to queries, keys, gated values, and level weights, then discards padded output rows. Padded source writes vanish, and padding occurs after the bucket's real tokens, so the inclusive mask on retained rows is unchanged. Thus padding preserves both the output and bucket summary in real arithmetic.

Since $0\le\widetilde m_\ell-m_\ell<C$, the total padded length obeys $\sum_\ell\widetilde m_\ell<T+CL_{\mathrm r}$. A bound for this schedule is
\begin{equation}
O\!\left((T+CL_{\mathrm r})[C(d_k+d_v)+d_kd_v]+Fd_kd_v\right),
\label{eq:reverse-appendix-padded-work}
\end{equation}
which retains the same asymptotic work at fixed $C$ and feature dimensions. The mathematical algorithm permits any positive integer $C$; our implementation restricts it to powers of two.

These definitions cover a bucket smaller than $C$, a partial final chunk, and a partially filled final bucket. When $N_\ell=1$, the exclusive incoming state is zero. When $e_\ell=T$, there are no later-query reads, even if the bucket has reached its nominal capacity. For $T=1$, the only chunk contains token zero, and the result is $\mu_0^{(0)}\eta_0\bv_0(\bk_0^\top\bq_0)$.

\paragraph{Persistent states and evaluation workspace.}
The reverse branch retains $L_{\mathrm r}$ state matrices for recurrent decoding, requiring $O(L_{\mathrm r}d_vd_k)$ storage per head. Training additionally requires sequence-level features and intermediate activations, whose memory usage depends on the implementation. The sequential-level algorithm accumulates into a single output buffer; evaluating levels concurrently may require additional temporary storage.

\paragraph{Recurrent continuation.}
At the end of prefill, $\Sbb_\ell=\Sbb_{T-1}^{\mathrm r,(\ell)}$ for every populated level. These summaries initialize Eq.~\eqref{eq:reverse-recurrence} without retaining individual attention keys or values. A partially filled final bucket continues to receive writes until its absolute-position boundary is reached; a newly reached bucket starts from zero. Continuation must preserve the absolute token position used by $\ell_{\mathrm r}(t)$ rather than restart bucket indexing at a new segment. Shared feature-generation history, where required, is separate from these reverse states.


\section{Synthetic Benchmark Details}
\label{app:mqar}

This appendix documents the MQAR experiment in Section~\ref{sec:synthetic-results}. The synthetic models are trained separately from the language-model checkpoints. The experiment contains 80 learning-rate tuning runs and 40 final runs.

\subsection{Data generation and scoring}
\label{app:mqar-data}

Our generator is adapted from the HazyResearch/Zoology \texttt{multiquery\_ar} task \citep{arora2024zoology}, with generator revision \texttt{1ad20d1}. It uses a vocabulary of 8192 synthetic token IDs and no text tokenizer. Each sequence has 512 input tokens and begins with $K$ distinct key--value pairs, each presented once. The remaining positions contain $K$ query keys among random filler tokens. We use one context pass and random token IDs at non-query positions. Query slots are sampled without replacement with weights proportional to $a j^{a-1}$ for one-based slot $j$, with $a=0.01$. This is the adopted generator's implementation default. At $K=128$, every available query slot is occupied; sampling changes which key is assigned to each query position.

For each $K\in\cK=\{32,64,96,128\}$, the training, validation, and test splits contain 100,000, 1000, and 1000 examples, respectively. This gives 400,000 training examples, 4000 validation examples, and 4000 test examples in total. Split base seeds are 123, 456, and 789. The resulting datasets are fixed and reused across all architectures and training seeds; the experiment manifest records derived seeds and SHA-256 hashes for all 12 data files.

Only value predictions at query positions are supervised. The remaining targets have ignore index $-100$. Targets are already aligned with their logits, so no additional target shift is applied. At evaluation, predictions are the argmax over vocabulary logits. For training seed $r$, let $\widehat y_{r,i,j}$ and $y_{i,j}$ be the predicted and target values for query $j$ of test example $i$. Accuracy for load $K$ and overall accuracy are
\[
A_r(K)=\frac{1}{N_K}\sum_{i=1}^{N_K}\frac{1}{K}
 \sum_{j=1}^{K}\mathbf1\{\widehat y_{r,i,j}=y_{i,j}\},
\qquad
A_r=\frac{1}{4}\sum_{K\in\cK}A_r(K),
\]
where $N_K=1000$. Thus the overall metric weights loads equally rather than weighting them by their different query counts. Tables and figures report $100A_r(K)$ or $100A_r$ in percent, aggregated over ten final training seeds using the arithmetic mean and sample standard deviation (denominator $10-1$). This variation is conditional on the shared fixed data; the standard deviations are not confidence intervals.

\subsection{Model configurations}
\label{app:mqar-models}

All four models have two layers, hidden size 64, tied input/output embeddings, initializer range 0.02, and no inference cache. All reported parameters are trainable, and tied embeddings are counted once. Table~\ref{tab:mqar-model-config} lists parameter counts and selected learning rates. We match depth and hidden width, while allowing the architecture-specific parameter counts and blocks below.

\begin{table}[!b]
\centering
\small
\begin{tabular*}{\linewidth}{@{\extracolsep{\fill}}lrr@{}}
\toprule
Model & Trainable parameters & Selected peak learning rate \\
\midrule
Mamba-2 & 580,656 & 0.00316227766 \\
Log-linear & 591,216 & 0.01 \\
Bilateral & 604,896 & 0.00316227766 \\
\arrayrulecolor{black!40}\cmidrule(lr){1-3}\arrayrulecolor{black}
Transformer & 655,680 & 0.01 \\
\bottomrule
\end{tabular*}
\caption{MQAR model sizes and selected learning rates. All models have two layers and hidden size 64. Counts include tied embeddings once.}
\label{tab:mqar-model-config}
\end{table}

Mamba-2, forward log-linear, and bilateral share state size 16, head dimension 16, expansion factor 2 (internal width 128 and eight heads), convolution width 4, one query/key group, and chunk size 64. They use the \texttt{chunk\_v2} implementation with their optional MLP disabled. The forward log-linear baseline is the standalone implementation with packed projections, corresponding to the experiment's \texttt{loglinear} artifact key.

Bilateral uses shared features, affine-free per-head branch RMS normalization with $\varepsilon=10^{-5}$, and learned two-way softmax mixing initialized to equal branch coefficients. Its reverse write gate is $2\sigma(\cdot)$ with zero initial weights and bias, giving multiplier one at initialization. 
Each reverse level read is additionally scaled by $2^{-\ell}$, giving the effective weight $2^{-\ell}\mu_t^{(\ell)}$, where $\mu_t^{(\ell)}$ is the learned weight in Eq.~\eqref{eq:bilateral-level-parameterization}. This fixed scaling uses the nominal full bucket size $2^\ell$, even when the bucket is partially filled. Both learned level-weight families use initial targets 1.0 before this scaling and input sensitivity 0.01. The local value feedthrough is added after branch fusion. The saved initial-probability field 0.9 applies to an alternative gate parameterization and is inactive for the twice-sigmoid gate used here.

The Transformer uses causal Llama-style attention with four query heads and four key/value heads, head dimension 16, a SwiGLU MLP of width 256, RoPE base 500,000, and RMS normalization epsilon $10^{-5}$. It uses PyTorch SDPA with the causal fast path and zero attention dropout. Its built-in SwiGLU block remains enabled, unlike the disabled optional MLPs of the Mamba-derived models.

\subsection{Learning-rate selection and training}
\label{app:mqar-training}

The training seed controls random parameter initialization and the order of training batches. It is separate from the data-generation seeds described in Appendix~\ref{app:mqar-data}; changing the training seed leaves the training, validation, and test datasets fixed.

For each architecture, we tune peak learning rate over
\[
\{10^{-3.5},10^{-3},10^{-2.5},10^{-2}\}
\]
using training seeds 0--4, giving $4\times4\times5=80$ tuning runs. Within each run, the checkpoint with highest validation accuracy is retained, with lower validation loss breaking accuracy ties. One learning rate per architecture is selected by highest mean validation accuracy across the five training seeds used for tuning. Remaining ties are broken by lower mean validation loss, fewer mean completed epochs, and then smaller learning rate. Test data are not used during tuning.

The selected rate is used for ten disjoint final training seeds, 10--19, giving 40 final runs. Training lasts at most 50 epochs and stops after an epoch whose validation accuracy is strictly greater than 0.99. Each final run restores its best-validation checkpoint and evaluates the test split once. Only completed runs enter the reported results.

We optimize query-only cross-entropy with AdamW and training/evaluation batch size 256. A batch contains examples from one load $K$; incomplete final batches are retained, and batch order is shuffled with the training seed. The schedule uses two warmup epochs followed by cosine decay toward learning rate $10^{-4}$ over the 50-epoch schedule. Weight decay is 0.1 for matrix-like parameters except explicit opt-outs, and zero for vectors and scalars. The global gradient-norm clipping threshold is $10^6$.

\begin{table}[tbp]
\centering
\small
\caption{MQAR test accuracy in percent, reported as mean $\pm$ sample standard deviation across ten training seeds (10--19). The overall score gives equal weight to the four memory loads. Standard deviations are in percentage points. Bold means indicate the highest accuracy among Mamba-2, log-linear, and bilateral.}
\label{tab:mqar-load-results}
\setlength{\tabcolsep}{4pt}
\begin{tabular*}{\linewidth}{@{\extracolsep{\fill}}lccccc@{}}
\toprule
Model & Overall & $K=32$ & $K=64$ & $K=96$ & $K=128$ \\
\midrule
Mamba-2 & $70.53\pm22.69$ & $90.72\pm21.61$ & $79.20\pm26.49$ & $63.67\pm24.67$ & $48.53\pm20.09$ \\
Log-linear & $76.37\pm14.35$ & $93.34\pm10.59$ & $83.89\pm15.70$ & $71.00\pm16.36$ & $57.25\pm15.23$ \\
Bilateral & $\mathbf{96.12}\pm3.79$ & $\mathbf{99.73}\pm0.38$ & $\mathbf{98.54}\pm1.95$ & $\mathbf{95.90}\pm5.10$ & $\mathbf{90.30}\pm7.85$ \\
\arrayrulecolor{black!40}\cmidrule(lr){1-6}\arrayrulecolor{black}
Transformer & $99.49\pm0.22$ & $99.64\pm0.13$ & $99.48\pm0.29$ & $99.40\pm0.30$ & $99.42\pm0.21$ \\
\bottomrule
\end{tabular*}
\end{table}

The overall experimental system contains eight NVIDIA RTX PRO 6000 Blackwell Server Edition GPUs; each MQAR run uses one visible GPU. The recorded environment is Python 3.11.15, PyTorch 2.12.1+cu130, and CUDA 13.0. Runs use float32 with TF32 enabled, high float32 matrix-multiplication precision, and \texttt{torch.compile}. Strict PyTorch deterministic algorithms are enabled with unsupported operations raising errors. cuDNN uses deterministic execution with benchmarking disabled and TF32 enabled; the cuBLAS workspace setting is \texttt{CUBLAS\_WORKSPACE\_CONFIG=:4096:8}.

\subsection{Complete test results}
\label{app:mqar-results}

Table~\ref{tab:mqar-load-results} gives the overall and per-load results underlying Figure~\ref{fig:mqar-results}. Transformer has the highest observed mean overall and at $K=64,96,128$; bilateral has the highest observed mean at $K=32$. These are comparisons of observed means, without a significance claim. Table~\ref{tab:mqar-seed-results} reports the overall test accuracy for each final training seed. In the main figure, bars and curves are means across training seeds, while error bars and shading show one sample standard deviation; shaded bands are clipped to the valid accuracy range.

\begin{table}[tbp]
\centering
\small
\caption{Overall MQAR test accuracy (\%) for all ten final training seeds. All runs use the same fixed training, validation, and test data. Bold denotes the best accuracy among Mamba-2, log-linear, and bilateral for each training seed.}
\label{tab:mqar-seed-results}
\begin{tabular*}{\linewidth}{@{\extracolsep{\fill}}rccc@{\hskip1.5em plus1fill}c@{}}
\toprule
Training seed & Mamba-2 & Log-linear & Bilateral & Transformer \\
\midrule
10 & 14.47 & 88.71 & \textbf{99.01} & 99.49 \\
11 & 78.13 & 81.78 & \textbf{96.86} & 99.17 \\
12 & 83.17 & 78.14 & \textbf{96.77} & 99.53 \\
13 & 69.52 & 45.54 & \textbf{87.38} & 99.59 \\
14 & 80.69 & 80.85 & \textbf{97.01} & 99.67 \\
15 & 83.68 & 80.84 & \textbf{98.99} & 99.25 \\
16 & 82.13 & 85.62 & \textbf{97.96} & 99.42 \\
17 & 78.30 & 54.62 & \textbf{97.73} & 99.67 \\
18 & 47.79 & 85.99 & \textbf{98.24} & 99.22 \\
19 & 87.44 & 81.60 & \textbf{91.22} & 99.86 \\
\bottomrule
\end{tabular*}
\end{table}

\section{Language Modeling: Evaluation Details and Full Results}
\label{app:language-modeling}

\subsection{Model configurations and evaluation protocol}
All selected language models use hidden width 1536, the tokenizer from \texttt{fla-hub/transformer-1.3B-100B}, training seed 1234, and a 6144-token training block. Results are from each run's final checkpoint, rather than a best-validation checkpoint, and are not averages across training seeds. Table~\ref{tab:lm-model-config} lists exact learnable parameter counts, including embeddings, MLPs, normalization parameters, and gates; tied input/output embeddings are counted once. The layer-matched comparison holds depth at 12, with different parameter counts. The approximately parameter-matched comparison varies depth and MLP configuration. The bilateral checkpoint, which uses shared features, normalized softmax fusion, and the reverse write gate, is reused in both panels of every results table.

As in the MQAR experiments, bilateral applies fixed reverse bucket-size scaling: each level-$\ell$ read uses the effective weight $2^{-\ell}\mu_t^{(\ell)}$, where $\mu_t^{(\ell)}$ is the learned weight in Eq.~\eqref{eq:bilateral-level-parameterization}. The scaling uses the nominal full bucket size $2^\ell$, even when the bucket is partially filled, and applies to the bilateral checkpoint in both matching settings.

\begin{table}[!b]
\centering
\small
\begin{tabular*}{\linewidth}{@{\extracolsep{\fill}}lrrl@{}}
\toprule
Model & Layers & Parameters & Configuration note \\
\midrule
\multicolumn{4}{l}{\emph{Layer-matched (12 layers)}} \\
\addlinespace[3pt]
Mamba-2 & 12 & 338,147,520 & One state matrix per head \\
Log-linear & 12 & 350,549,952 & Forward hierarchy \\
Bilateral & 12 & 365,608,320 & Shared, normalized, gated \\
\arrayrulecolor{black!40}\cmidrule(lr){1-4}\arrayrulecolor{black}
Transformer & 12 & 275,682,816 & MLP width 2048 \\
\midrule
\multicolumn{4}{l}{\emph{Approximately parameter-matched}} \\
\addlinespace[3pt]
Mamba-2 & 13 & 362,230,352 & One state matrix per head \\
Log-linear & 12 & 365,590,464 & Forward hierarchy; MLP width 2320 \\
Bilateral & 12 & 365,608,320 & Same bilateral checkpoint \\
\arrayrulecolor{black!40}\cmidrule(lr){1-4}\arrayrulecolor{black}
Transformer & 17 & 370,070,016 & 24 query/KV heads; head width 64 \\
\bottomrule
\end{tabular*}
\caption{Selected language-model configurations. Counts include all learnable parameters. The approximately parameter-matched log-linear model uses an MLP width of 2320. Both panels reuse the same bilateral checkpoint.}
\label{tab:lm-model-config}
\end{table}

\paragraph{Pretraining data and optimization.}
All models use the same prepared DHSA Long Data Collections corpus and overall pretraining protocol. The corpus is split into 99\% training data and 1\% validation data. The prepared training split contains 2,180,019 retained chunks and 10,000,808,832 valid next-token prediction targets; training covers one complete epoch of this split. Sequences contain at most 6144 tokens, with trailing chunks retained when they contain at least 512 tokens. Padding targets are excluded from the loss.

For bilateral, we use AdamW with learning rate $3\times10^{-4}$, weight decay 0.1 on eligible matrix-like parameters, and global gradient-norm clipping at 1.0. Biases, normalization parameters, and designated dynamics and level-routing parameters are excluded from weight decay. Other AdamW options use the PyTorch defaults. The learning rate warms up for 1700 optimizer updates and then follows cosine decay to $3\times10^{-5}$. Training uses eight GPUs, one sequence per GPU per microbatch, and two gradient-accumulation steps, giving an effective batch of 16 sequences per optimizer update and approximately 136,252 updates over one epoch. Parameters and optimizer state remain in float32, with bfloat16 autocast for mixed-precision computation. TF32 and \texttt{torch.compile} in its default mode are enabled; gradient checkpointing is disabled. The next-token loss is normalized by the total number of valid prediction targets across GPUs and accumulation steps, so shorter retained chunks contribute in proportion to their valid target counts.

\paragraph{Standard language-model evaluation.}
WikiText-2 rolling perplexity uses a 6144-token context for all seven selected models, matching the training block length. Every evaluation includes all 62 documents, without an example limit, using batch size 16 per GPU, bfloat16, and evaluation seed 42. Document, prompt, and target hashes match across all seven evaluations. The five non-Transformer configurations use separate 6144-token reevaluations of the final checkpoints; both Transformers retain their original 6144-token evaluations. For LAMBADA and the zero-shot accuracy tasks, the original evaluation limits remain 6144 for Transformers and 8192 for the other models. These tasks fit within both limits; the longest recorded LAMBADA input is 230 tokens. We retain the selected metrics used in the Log-Linear Attention comparison: \texttt{acc} for LAMBADA, PIQA, WinoGrande, and ARC-Easy, and \texttt{acc\_norm} for HellaSwag and ARC-Challenge. The LM-Eval average is the unweighted mean of these six scores and excludes perplexity.

\paragraph{RULER-NIAH.}
We use the same prepared examples for all models, with 500 examples in each task--length cell at 2048, 4096, and 6144 tokens. All examples are retained. Following \citet{guo2026loglinear}, the task labels are S-NIAH-1 (pass-key retrieval), S-NIAH-2 (number in haystack), S-NIAH-3 (uuid in haystack), MK-NIAH-1 (multi-key line retrieval), MQ-NIAH (multi-query), and MV-NIAH (multi-value). Averages are taken over these six tasks, rather than the full RULER suite. All three lengths are within the training block length.

\paragraph{Downstream retrieval.}
The five full-dataset tasks are SWDE, SQuAD completion, FDA, TriviaQA, and DROP; the remaining tasks are NQ-512, NQ-1024, and NQ-2048. We average the eight task scores with equal weight. All models use an 8192-token retrieval limit. For both Transformers, this is an evaluation override with standard, unscaled RoPE extrapolation beyond the trained 6144-token context; saved training configurations and model weights are unchanged. The other models evaluate at 8192 without context extension. Prompts, scoring, retained dataset fingerprints, and NQ exclusions match across the selected runs. NQ-2048 excludes the same six overlength rows for every model, retaining 3151 of 3157 examples; all other tasks retain all examples. The context override and exclusions are recorded in the affected evaluation cells' metadata.

\begin{table}[!b]
\centering
\small
\setlength{\tabcolsep}{2.5pt}
\begin{tabular*}{\linewidth}{@{\extracolsep{\fill}}lrrrrrrrrr@{}}
\toprule
Model & Wiki. PPL & Lamb. PPL & Lamb. & PIQA & Hella. & Wino. & ARC-E & ARC-C & Avg. \\
\midrule
\multicolumn{10}{l}{\emph{Layer-matched (12 layers)}} \\
\addlinespace[3pt]
Mamba-2 & 31.37 & 39.66 & 27.93 & 63.71 & 33.83 & 50.20 & 45.71 & 24.15 & 40.92 \\
Log-linear & 30.23 & 33.78 & 30.53 & \textbf{64.20} & 34.13 & 51.38 & \textbf{46.97} & \textbf{26.19} & \textbf{42.23} \\
Bilateral & \textbf{30.21} & \textbf{33.20} & \textbf{31.13} & 62.79 & \textbf{34.15} & \textbf{51.46} & 45.92 & 23.63 & 41.51 \\
\arrayrulecolor{black!40}\cmidrule(lr){1-10}\arrayrulecolor{black}
Transformer & 32.53 & 40.75 & 30.56 & 61.70 & 31.64 & 52.01 & 44.57 & 24.40 & 40.81 \\
\midrule
\multicolumn{10}{l}{\emph{Approximately parameter-matched}} \\
\addlinespace[3pt]
Mamba-2 & 30.93 & 33.97 & 30.37 & 63.00 & 33.85 & 51.62 & 45.79 & \textbf{24.91} & \textbf{41.59} \\
Log-linear & 30.38 & 37.04 & 29.23 & \textbf{63.66} & 33.96 & \textbf{52.17} & 44.57 & 23.98 & 41.26 \\
Bilateral & \textbf{30.21} & \textbf{33.20} & \textbf{31.13} & 62.79 & \textbf{34.15} & 51.46 & \textbf{45.92} & 23.63 & 41.51 \\
\arrayrulecolor{black!40}\cmidrule(lr){1-10}\arrayrulecolor{black}
Transformer & 30.45 & 35.14 & 31.94 & 63.87 & 33.28 & 52.72 & 44.82 & 24.15 & 41.80 \\
\bottomrule
\end{tabular*}
\caption{Language-modeling and commonsense-reasoning results. The first two metric columns are perplexities ($\downarrow$); the remaining columns are accuracy percentages ($\uparrow$). The six-task average excludes perplexities. WikiText-2 uses a 6144-token context for all models. Bold denotes the best result among Mamba-2, log-linear, and bilateral within each panel.}
\label{tab:lm-selected-task-results}
\end{table}

\subsection{Standard language modeling and commonsense reasoning}
Table~\ref{tab:lm-selected-task-results} reports the two perplexities and all six selected task scores. Perplexities are unscaled; accuracy values and their average are percentages. Abbreviations in the table are Wiki. (WikiText-2), Lamb. (LAMBADA), Hella. (HellaSwag), Wino. (WinoGrande), ARC-E (ARC-Easy), and ARC-C (ARC-Challenge). HellaSwag and ARC-Challenge use length-normalized accuracy; the other four tasks use accuracy. Bold denotes the best result among Mamba-2, log-linear, and bilateral within each panel. 

\subsection{RULER-NIAH task-level results}
Tables~\ref{tab:ruler-niah-2048}--\ref{tab:ruler-niah-6144} give all six task scores at each evaluated length. Scores and their unweighted averages are percentages; higher is better. Bold denotes the best result among Mamba-2, log-linear, and bilateral within each panel.

\begin{table}[tbp]
\centering
\caption{RULER-NIAH results at context length 2048. Each task uses the same 500 prepared examples for every model. Scores are percentages; higher is better. Avg. is the unweighted mean over the six tasks. Bold denotes the best result among Mamba-2, log-linear, and bilateral within each panel.}
\label{tab:ruler-niah-2048}
\small
\setlength{\tabcolsep}{3pt}
\begin{tabular*}{\linewidth}{@{\extracolsep{\fill}}lrrrrrrr@{}}
\toprule
Model & S-NIAH-1 & S-NIAH-2 & S-NIAH-3 & MK-NIAH-1 & MQ-NIAH & MV-NIAH & Avg. \\
\midrule
\multicolumn{8}{l}{\emph{Layer-matched (12 layers)}} \\
\addlinespace[3pt]
Mamba-2 & 77.80 & 13.40 & 33.60 & 26.60 & 19.15 & 17.65 & 31.37 \\
Log-linear & 99.80 & 68.20 & 11.20 & 34.40 & 24.05 & 24.25 & 43.65 \\
Bilateral & \textbf{100.00} & \textbf{100.00} & \textbf{70.60} & \textbf{49.60} & \textbf{28.80} & \textbf{30.10} & \textbf{63.18} \\
\arrayrulecolor{black!40}\cmidrule(lr){1-8}\arrayrulecolor{black}
Transformer & 99.80 & 100.00 & 65.60 & 59.20 & 39.45 & 40.45 & 67.42 \\
\midrule
\multicolumn{8}{l}{\emph{Approximately parameter-matched}} \\
\addlinespace[3pt]
Mamba-2 & 99.60 & 98.60 & 1.00 & 26.00 & 20.20 & 16.30 & 43.62 \\
Log-linear & 99.00 & 98.40 & 70.00 & 30.20 & 25.30 & 25.45 & 58.06 \\
Bilateral & \textbf{100.00} & \textbf{100.00} & \textbf{70.60} & \textbf{49.60} & \textbf{28.80} & \textbf{30.10} & \textbf{63.18} \\
\arrayrulecolor{black!40}\cmidrule(lr){1-8}\arrayrulecolor{black}
Transformer & 100.00 & 99.80 & 86.00 & 69.80 & 45.60 & 35.15 & 72.72 \\
\bottomrule
\end{tabular*}
\end{table}

\begin{table}[!b]
\centering
\caption{RULER-NIAH results at context length 4096. Each task uses the same 500 prepared examples for every model. Scores are percentages; higher is better. Avg. is the unweighted mean over the six tasks. Bold denotes the best result among Mamba-2, log-linear, and bilateral within each panel.}
\label{tab:ruler-niah-4096}
\small
\setlength{\tabcolsep}{3pt}
\begin{tabular*}{\linewidth}{@{\extracolsep{\fill}}lrrrrrrr@{}}
\toprule
Model & S-NIAH-1 & S-NIAH-2 & S-NIAH-3 & MK-NIAH-1 & MQ-NIAH & MV-NIAH & Avg. \\
\midrule
\multicolumn{8}{l}{\emph{Layer-matched (12 layers)}} \\
\addlinespace[3pt]
Mamba-2 & 58.40 & 41.80 & 1.80 & 22.20 & 20.20 & 19.00 & 27.23 \\
Log-linear & 99.40 & 62.40 & 8.00 & 30.20 & 21.20 & 21.95 & 40.52 \\
Bilateral & \textbf{99.80} & \textbf{98.60} & \textbf{55.40} & \textbf{49.60} & \textbf{23.65} & \textbf{27.80} & \textbf{59.14} \\
\arrayrulecolor{black!40}\cmidrule(lr){1-8}\arrayrulecolor{black}
Transformer & 98.40 & 100.00 & 47.80 & 64.80 & 29.80 & 26.90 & 61.28 \\
\midrule
\multicolumn{8}{l}{\emph{Approximately parameter-matched}} \\
\addlinespace[3pt]
Mamba-2 & 99.00 & 42.00 & 0.00 & 17.60 & 14.20 & 8.10 & 30.15 \\
Log-linear & 94.00 & 82.60 & 17.80 & 29.00 & 21.75 & 22.45 & 44.60 \\
Bilateral & \textbf{99.80} & \textbf{98.60} & \textbf{55.40} & \textbf{49.60} & \textbf{23.65} & \textbf{27.80} & \textbf{59.14} \\
\arrayrulecolor{black!40}\cmidrule(lr){1-8}\arrayrulecolor{black}
Transformer & 100.00 & 99.80 & 73.20 & 64.40 & 28.65 & 25.75 & 65.30 \\
\bottomrule
\end{tabular*}
\end{table}

\begin{table}[H]
\centering
\caption{RULER-NIAH results at context length 6144. Each task uses the same 500 prepared examples for every model. Scores are percentages; higher is better. Avg. is the unweighted mean over the six tasks. Bold denotes the best result among Mamba-2, log-linear, and bilateral within each panel.}
\label{tab:ruler-niah-6144}
\small
\setlength{\tabcolsep}{3pt}
\begin{tabular*}{\linewidth}{@{\extracolsep{\fill}}lrrrrrrr@{}}
\toprule
Model & S-NIAH-1 & S-NIAH-2 & S-NIAH-3 & MK-NIAH-1 & MQ-NIAH & MV-NIAH & Avg. \\
\midrule
\multicolumn{8}{l}{\emph{Layer-matched (12 layers)}} \\
\addlinespace[3pt]
Mamba-2 & 38.60 & 6.60 & 6.20 & 12.00 & 5.15 & 8.20 & 12.79 \\
Log-linear & 94.80 & 15.00 & 5.60 & 17.40 & 15.30 & 14.15 & 27.04 \\
Bilateral & \textbf{99.80} & \textbf{89.80} & \textbf{23.60} & \textbf{45.80} & \textbf{20.90} & \textbf{20.25} & \textbf{50.03} \\
\arrayrulecolor{black!40}\cmidrule(lr){1-8}\arrayrulecolor{black}
Transformer & 100.00 & 100.00 & 42.80 & 60.40 & 27.80 & 27.65 & 59.77 \\
\midrule
\multicolumn{8}{l}{\emph{Approximately parameter-matched}} \\
\addlinespace[3pt]
Mamba-2 & 97.40 & 4.20 & 0.80 & 2.20 & 3.55 & 2.50 & 18.44 \\
Log-linear & 86.40 & 32.00 & 5.00 & 9.80 & 13.80 & 14.85 & 26.97 \\
Bilateral & \textbf{99.80} & \textbf{89.80} & \textbf{23.60} & \textbf{45.80} & \textbf{20.90} & \textbf{20.25} & \textbf{50.03} \\
\arrayrulecolor{black!40}\cmidrule(lr){1-8}\arrayrulecolor{black}
Transformer & 100.00 & 91.60 & 57.00 & 56.60 & 27.25 & 20.00 & 58.74 \\
\bottomrule
\end{tabular*}
\end{table}

\subsection{Downstream retrieval task-level results}
Table~\ref{tab:retrieval-condition-results} reports all eight retrieval tasks. Scores and their unweighted average are percentages; higher is better. ``SQuAD'' denotes SQuAD completion. The NQ suffixes identify the three evaluated NQ variants. Bold denotes the best result among Mamba-2, log-linear, and bilateral within each panel.

\begin{table}[H]
\centering
\small
\setlength{\tabcolsep}{2.5pt}
\begin{tabular*}{\linewidth}{@{\extracolsep{\fill}}lrrrrrrrrr@{}}
\toprule
Model & SWDE & SQuAD & FDA & TriviaQA & DROP & NQ-512 & NQ-1024 & NQ-2048 & Avg. \\
\midrule
\multicolumn{10}{l}{\emph{Layer-matched (12 layers)}} \\
\addlinespace[3pt]
Mamba-2 & 20.61 & 28.18 & 10.62 & 38.86 & 16.05 & 13.78 & 11.53 & 10.44 & 18.76 \\
Log-linear & 24.93 & 31.20 & 15.43 & \textbf{39.75} & \textbf{17.87} & 16.63 & 13.62 & 12.28 & 21.46 \\
Bilateral & \textbf{29.79} & \textbf{33.51} & \textbf{32.76} & 39.57 & 17.44 & \textbf{17.83} & \textbf{15.93} & \textbf{14.47} & \textbf{25.16} \\
\arrayrulecolor{black!40}\cmidrule(lr){1-10}\arrayrulecolor{black}
Transformer & 37.44 & 33.71 & 62.07 & 39.10 & 17.68 & 18.06 & 16.06 & 13.93 & 29.76 \\
\midrule
\multicolumn{10}{l}{\emph{Approximately parameter-matched}} \\
\addlinespace[3pt]
Mamba-2 & 20.16 & 30.70 & 13.88 & 39.04 & \textbf{18.54} & 16.12 & 13.21 & 11.14 & 20.35 \\
Log-linear & 22.95 & 30.23 & 18.60 & 38.21 & 18.30 & 17.48 & 14.89 & 12.38 & 21.63 \\
Bilateral & \textbf{29.79} & \textbf{33.51} & \textbf{32.76} & \textbf{39.57} & 17.44 & \textbf{17.83} & \textbf{15.93} & \textbf{14.47} & \textbf{25.16} \\
\arrayrulecolor{black!40}\cmidrule(lr){1-10}\arrayrulecolor{black}
Transformer & 43.11 & 33.38 & 67.33 & 40.17 & 17.20 & 19.99 & 18.50 & 16.95 & 32.08 \\
\bottomrule
\end{tabular*}
\caption{Downstream retrieval scores under the common evaluation protocol. SWDE, SQuAD completion, FDA, TriviaQA, and DROP use their full-dataset evaluations. Scores are percentages; higher is better. Avg. is the unweighted mean over the eight retrieval tasks, counting each NQ variant separately. Bold denotes the best result among Mamba-2, log-linear, and bilateral within each panel.}
\label{tab:retrieval-condition-results}
\end{table}

\end{document}